# A Review on Automated Brain Tumor Detection and Segmentation from MRI of Brain

Sudipta Roy[1], Sanjay Nag[2], Indra Kanta Maitra[3], Prof. Samir Kumar Bandyopadhyay[4]

[1] Department of Computer Science and Engineering, University of Calcutta,
92 A.P.C. Road, Kolkata-700009, India.
[3] Research Scholar, Department of Computer Science & Engineering,
University of Calcutta, India.
[2] Research Fellow, Department of Computer Science & Engineering,
University of Calcutta, India.
[4] Vice Chancellor, West Bengal University of Technology
Kolkata, West Bengal, India.

sudiptaroy01@yahoo.com, sanjaynag75@gmail.com, ikm.1975@yahoo.com, skb1@vsnl.com

**ABSTRACT:** Tumor segmentation from magnetic resonance imaging (MRI) data is an important but time consuming manual task performed by medical experts. Automating this process is a challenging task because of the high diversity in the appearance of tumor tissues among different patients and in many cases similarity with the normal tissues. MRI is an advanced medical imaging technique providing rich information about the human soft-tissue anatomy. There are different brain tumor detection and segmentation methods to detect and segment a brain tumor from MRI images. These detection and segmentation approaches are reviewed with an importance placed on enlightening the advantages and drawbacks of these methods for brain tumor detection and segmentation. The use of MRI image detection and segmentation in different procedures are also described. Here a brief review of different segmentation for detection of brain tumor from MRI of brain has been discussed.

**KEYWORDS:** *MRI of Brain, Tumor Segmentation, Tumor Detection, Automated System, Pre-processing, Filtering.*

**1. INTRODUCTION**: Magnetic resonance imaging of brain image computing has very increased field of medicine by providing some different methods to extract and visualize information from medical data, acquired using various acquisition modalities. Brain tumor segmentation is a significant process to extract information from complex MRI of brain images. Diagnostic imaging is a very useful tool in medical today. Magnetic resonance imaging (MRI), computed tomography (CT), digital mammography, and other imaging processes give an efficient means for detecting different type of diseases. The automated detection methodology have deeply improved knowledge of normal and diseased examination for medical research and are a important part in diagnosis and treatment planning when the number of patients increases[1]. Segmentation has spacious application in medical imaging field such as MRI of brain, MRI of human knee, etc. for analyzing MRI of brain, anatomical structures such as bones, muscles blood vessels, tissue types, pathological regions such as cancer, multiple sclerosis lesions and for dividing an entire image into sub regions such as the white matter (WM), gray matter (GM) and cerebrospinal fluid (CSF) spaces of the brain automated delineation of different image components are used. Thus in the field of MRI of brain tumor segmentation from brain image is significant as MRI is particularly suitable for brain studies because of its excellent contrast of soft issues, non invasive characteristic and a high spatial resolution. Brain tumor segmentation partitions a portion into mutually special and pooped regions such that each region of interest is spatially contiguous and the pixels within the region are homogeneous with respect to a predefined criterion. Mostly, homogeneity conditions include values of concentration, texture, color, range, surface normal and surface curvatures. Through past many researchers have prepared important research in the field of brain tumor segmentation but still now it is very important research fields. Medical history, biopsy–whereby a small amount of brain tissue is excised and analyzed under the microscope–and imaging studies are all important to reach a diagnosis of brain tumor. Standard x-rays and computed tomography (CT) can initially be used in the diagnostic process. However, MRI is generally more useful because it provides more detailed information about tumor type, position and size. For this reason, MRI is the imaging study of choice for the diagnostic work up and, thereafter, for surgery and monitoring treatment outcomes [2]. Thus here short introduction with MRI, brain tumor and automated system also discussed.

Rest of the part organized as follows: Section 2 describes the Magnetic resonance imaging of Brain Image. In Section 3 there is a description about Brain Tumor. Section 4 describes the automated system and Section 5 consists of Preprocessing. Colour Fundamentals is described in Section 6 and Section 7 consists of Segmentation. In Section 8 there are Summery and Conclusion. Section 9 consists of Reference.

## 2. Magnetic resonance imaging (MRI) :
A magnetic resonance imaging instrument or MRI Scanner [3] uses powerful magnets to polarize and excite hydrogen nuclei i.e. proton in water molecules in human tissue, producing a detectable signal which is spatially encoded, resulting in images of the body [4]. MRI mainly uses three electromagnetic fields they are : i) A very strong static magnetic field to polarize the hydrogen nuclei, named as the static field, ii) A weaker time varying field(s) for spatial encoding, named as the gradient field, iii) A weak radio frequency field for manipulation of hydrogen nuclei to produce measurable signals collected through RF antenna. The variable behaviour of protons within different tissues leads to differences in tissue appearance. The different positioning of MRI of brain with T1 and T2 weight is shown below.

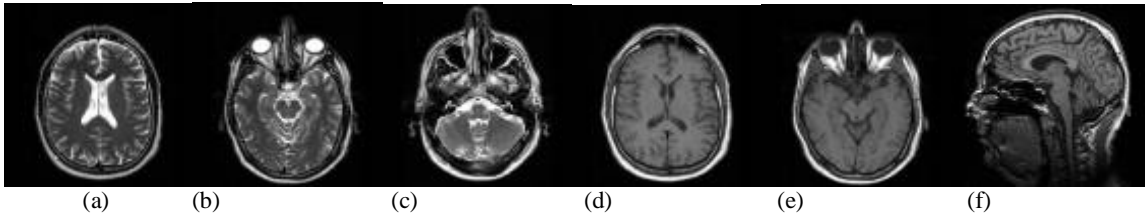

(a)　　(b)　　(c)　　(d)　　(e)　　(f)

*Figure 1* : MRI of brain cited by http://www.mr-tip.com/serv1.php?type=isimg. T2 weighted MR image (a) brain shows cortex, lateral ventricle, and falx cerebri, (b) brain shows eyeballs with optic nerve, medulla, vermis, and temporal lobes with hippocampal regions, (c) head shows maxillary sinus, nasal septum, clivus, inner ear, medulla, and cerebellum. T1 weighted MR image (d) brain shows cortex, white and grey matter, third and lateral venticles, putamen, frontal sinus and superior sagittal sinus, (e) brain shows eyeballs with optic nerve, medulla, vermis, and temporal lobes with hippocampal regions,(f) brain shows cortex with white and grey matter, corpus callosum, lateral ventricle, thalamus, pons and cerebellum from the same patients

## 3. Brain Tumor:
A brain tumor is a mass of cells that have grown and multiplied uncontrollable i.e. a brain tumor is an uncontrolled growth of solid mass formed by undesired cells either normally found in the different part of the brain such as glial cells, neurons, lymphatic tissue, blood vessels, pituitary and pineal gland, skull, or spread from cancers mainly located in other organs [5]. Brain tumors are classified based on the type of tissue involved in the brain, the positioning of the tumor in the brain, whether it is benign tumor or malignant tumor and other different considerations. Brains tumors are the solid portion permeate the surrounding tissues or distort the surrounding structures [6]. There are different type of brain tumor they are i) Gliomas, ii) Medulloblastoma, iii) Lymphoma, iv) Meningioma, v) Craniopharyngioma, vi) Pituitary adenoma.

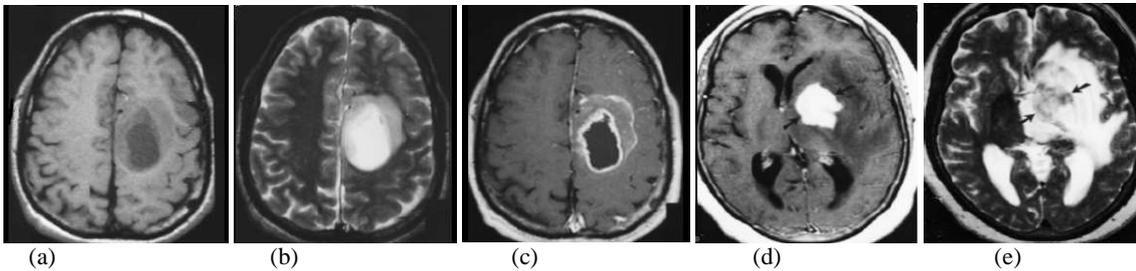

(a)　　(b)　　(c)　　(d)　　(e)

*Figure 2*: A set of brain tumor images from MRI of brain output cited by Herbert H. Engelhard et al.(2003)[7]. a) Axial T1-weighted with tumor, b) T2-weighted with central positioning tumor, c) Contrast enhanced T1-weighted image showing ring formed tumor, d) Contrast enhanced T1-weighted image with high grade oligodendro glioma e) T2-weighted image with high grade oligodendro glioma from the same patient.

## 4. Automated System:
Automated system (detection) of brain tumor through MRI is basically called Computer-Aided Diagnosis (CAD) system. The CAD system can provide highly accurate reconstruction of the original image i.e. the valuable outlook and accuracy of earlier brain tumor detection. It consists of two or more stage. In the initial stage pre-processing has required after that stages post-processing i.e. segmentation are required. Then detection strategies and other information, feature extraction, feature selection, classification, and performance analysis are compared and studied. Pre-processing techniques are used to improvement of image quality and remove small artefacts and noise for the accurate detection of the undesired regions in MRI. Post-processing is used to segment with different strategy the brain tumor from the MRI of brain images.

In this review, here focus on the appearance of tumors in MRI images, the grade of tumors and some general information which will be useful in the detection, segmentation and interpretation of brain tumors from MRI images. An automated brain tumor detection procedure follows some steps which is shown diagram below.

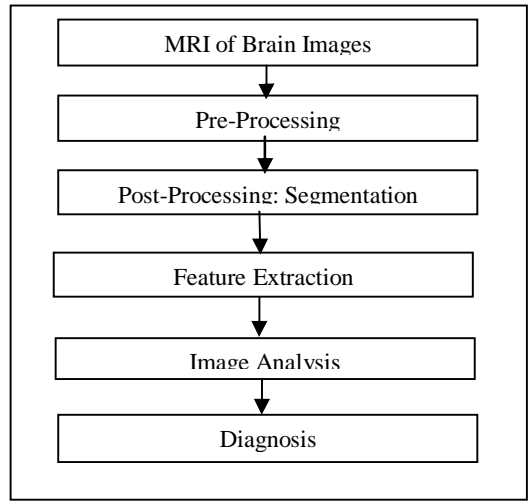

**Figure 3**: Brain tumor detection steps

**5. PREPROCESSING:** Pre-processing mainly involves those operations that are normally necessarily prior to the main goal analysis and extraction of the desired information and normally geometric corrections of the original actual image. These improvements include correcting the data for irregularities and unwanted atmospheric noise, removal of non-brain element image and converting the data so they correctly reflected in the original image. Segmentation is the process of partitioning an image to several segments but the main difficulties in segmenting an images are i) Noise, ii) Blur Low Contrast, iii) The bias field (the occurrence of smoothly varying intensities within tissues), iv) The partial-volume effect (a voxel contributes in multiple tissue types). Image filtering and enhancement stage is the most obvious part of medical image processing. This pre-processing stage is used for reducing image noise, highlighting important portions, or displaying obvious portions of digital images [8]. Some more techniques can employ medical image processing of coherent echo signals prior to image generation and some of the images are hanging from clip hence they may produce noise. The enhancement stage includes resolution enhancement; contrast enhancement. These are used to suppress noise and imaging of spectral parameters. After this stage the medical image is converted into standard image without noise, film artefacts and labels.

*5.1 Existing de-noising methods*: In spite of the presence of substantial number of state of the art methods of de-noising but accurate removal of noise from MRI image is a challenge. Methods such as use of standard filters to more advanced filters, nonlinear filtering methods, anisotropic nonlinear diffusion filtering, a Markov random field (MRF) models, wavelet models, non-local means models (NL-means) and analytically correction schemes. These methods are almost same in terms of computation cost, de-noising, quality of de-noising and boundary preserving. So, de-noising is still an open issue and de-noising methods needs improvement. Linear filters reduce noise by updating pixel value by weighted average of neighbourhood but degrade the image quality substantially. On the other hand, non linear filters preserve edges but degrade fine structures.

i) *Non-local (NL)*: This method exploits the redundant information in images [9]. The pixel values are replacement by taking weighted average of locality similar to the neighbourhood surrounding of the image. MRI images, consists of non-repeated details due to noise, complex structures, smear in acquisition and the partial volume effect originating from the low sensor resolution that is abolished by this type of method.

ii) *Analytical correction method*: This method attempts to estimate which is noise and which is noise-free signal from observed image. This method uses maximum likelihood estimation i.e. depending upon the signal [10] to estimate noise and noise free portion of images. Neighbourhood smoothing is used to estimate noise free image by considering signal in small region to be constant. Edges in the image are degraded.

iii) *A Markov random field method (MRF)*: In this method spatial correlation information is used to preserve fine detail [11], i.e., spatial regularization of the noise estimation is performed. In MRF method, the updating of pixel value is done by iterated conditional modes and simulated annealing with maximizing a posterior estimate.

iv) *Wavelet based methods*: In frequency domain these method is used for de noising and preserving the actual signal. Application of wavelet based methods on MRI images makes the wavelet and scaling coefficients biased. This problem is solved by squaring the MRI image by non central chi-square distribution method [12]. These make the scaling coefficients independent of the signal and thus can be easily removed [13]. In case of low SNR images, finer details are not preserved [14].

These are the major techniques used in filtering image and to enhance the image quality use filtering with some other procedure which is describe in the next section.

**5.2 Images Enhancement and Filtering**: Image enhancement is the improvement of digital image quality without any knowledge about the original source image degradation. The enhancement methods mainly divide into two methods, direct and indirect methods. In direct method is to show the contrast of the image and then enhance the contract but in the indirect method contrast of the image is not essential. Under-enhanced when some regions of the image may be over-enhanced are the great disadvantage of the contrast enhancement methods. Mainly image enhancements are the intensity and contrast manipulation, noise reduction, undesirable background removal, edges Sharpening, filtering etc. Image enhancement methods improve the visual appearance of images from MRI and the contrast enhancing brain volumes were linearly associated. The enhancement activities are removal of film artefacts and labels, filtering the images. Median Filter, Low pass Filter, Gradient Based Method, Prewitt edge-finding filter, Nonlinear Filter, V-filter, and other filter with contrast Enhanced filter are shortly describe below.

There are several filtering technology's which can improve the MRI image quality but there are several advantage and disadvantage which describe very shortly individually.

**5.2.1 Median Filter:** Median Filter remove the noise with high frequency components from MRI without disturbing the edges and it is used to reduce' salt and pepper' noise. This technique calculates the median values i.e. set median value pixels values of the surrounding pixels to determine the new de-noised value of the pixel [15]. A median is calculated by sorting all pixel values by their size, then selecting the median value as the new value for the pixel. The basic function for median image is written below in equation (1), where f(x,y) output median and g(x,y) is the original values.

$$\hat{f}(x, y) = \underset{(i, j) \in N}{median} \{g(i, j)\}$$

Before filtering if the intensity values are 46, 58, 47, 49, 41, 45, 42, 55, and 59. So the ascending orders of the pixel intensity values are 41, 42, 45, 46, **47**, 49, 55, 58, and 59. So the median value is 47 and this 47 replace the value 41 in the actual output of the median filter.

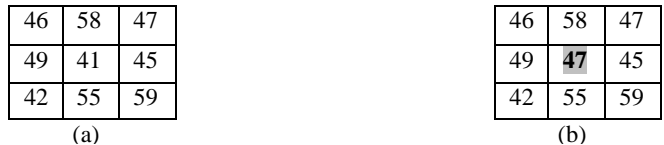

| 46 | 58 | 47 |
|---|---|---|
| 49 | 41 | 45 |
| 42 | 55 | 59 |
(a)

| 46 | 58 | 47 |
|---|---|---|
| 49 | **47** | 45 |
| 42 | 55 | 59 |
(b)

*Figure 4: Median Filter a) Shows pixel intensity value before filtering and b) Shows pixel intensity value after filtering.*

Different type of median filter are shortly describe below

    **i)** *Min-Max Median Filter:* It is conditional nonlinear filter with (3x3) window is use for scanning the image left to right and top to bottom. In this methods center pixel of window (2, 2) is considered as a test pixel and center pixel value is replaced by the median value of the neighbour pixel which pixel is less than the minimum value and greater than the maximum value [16]. Then center pixel is considered as corrupted pixel that is why value is replaced by median value of pixels nearby in window otherwise pixel is treated as non corrupted pixel and kept pixel value remain unchanged.

    *ii) Center Weighted Median Filter*: It is an extension with weighted median filter in which it produce more weight to the center. It is very much effective to smooth the image with capability of image enhancement [17]. In this procedure center pixel of (2n+1) square window measured as test pixel and if center pixel (n+1,n+1) less than minimum value present in rest of pixel in window and greater than maximum value present in rest of pixel in window then center pixel is treated as corrupted pixel.

    *iii) Adaptive Median Filter*: It is conditional non-linear filter with two conditions that are used to detect corrupted pixels and to verify correctness of median value, and also uses increasing shifting windows size until correct value of median is calculated and corrected median value replace the corrupted pixel to reduce noise [18]. If the median value or test pixel is within minimum and maximum limit then it is corrected pixel otherwise treated as corrupted pixel and then increase the window size until correct median value within the limit is produce, then the corrupted value is replaced by median value.

    iv) Progressive Switching Median Filter: It is [19] is conducted by two step process in which first step is noise pixel identification with fixed size window size. If the test pixel is not within maximum or minimum value present rest of the pixel then this pixel is treated as corrupted pixel and in the second step these corrupted pixel are replaced by the median value intensity. Median value out of maximum/minimum limit is treated as corrupted median values which are not used and then window size increases. Here median value is calculated same as in adaptive median filter without considering the corrupted pixel present in window.

As low frequency image is generated and the images are enhanced using median filter, suppresses isolated noise without blurring sharp edges. From figure 5 we see that median filter produce very good results and the median is much less sensitive than the mean to extreme values thus median filtering is therefore better able to remove this outlier without reducing the sharpness of the image.

**5.2.2 Mean Filter:** In case of mean or average filter replace the center of the pixel by mean (average) value of the pixels. Thus if a square window of size 2k+1 is used, where k within 1 to n, odd width and height are taking and average value is replaced by (k+1, k+1) position [20,21,24,25]. The basic function for median is written below in equation (2), where f(x,y) output mean and g(x,y) is the original values.

$$f(x,y) = \frac{1}{K} \sum_{i=1}^{K} g_i(x,y)$$

Each time of scan value of central pixel of window is replaced by the average value of its neighbouring pixels comes within the window. If k=1 the window size is 3×3 and center position is 2×2 is shown in figure 5 Before filtering if the intensity value are 46, 58, 47,49, 41, 45, 42, 55, 58 . So the average value is 49 and then 41 are replaced by 49.

| 46 | 58 | 47 |
|----|----|----|
| 49 | 41 | 45 |
| 42 | 55 | 58 |
(a)

| 46 | 58 | 47 |
|----|----|----|
| 49 | **49** | 45 |
| 42 | 55 | 58 |
(b)

*Figure 5* : Mean filter: a) Shows pixel intensity value before filtering and b) Shows pixel intensity value after filtering.

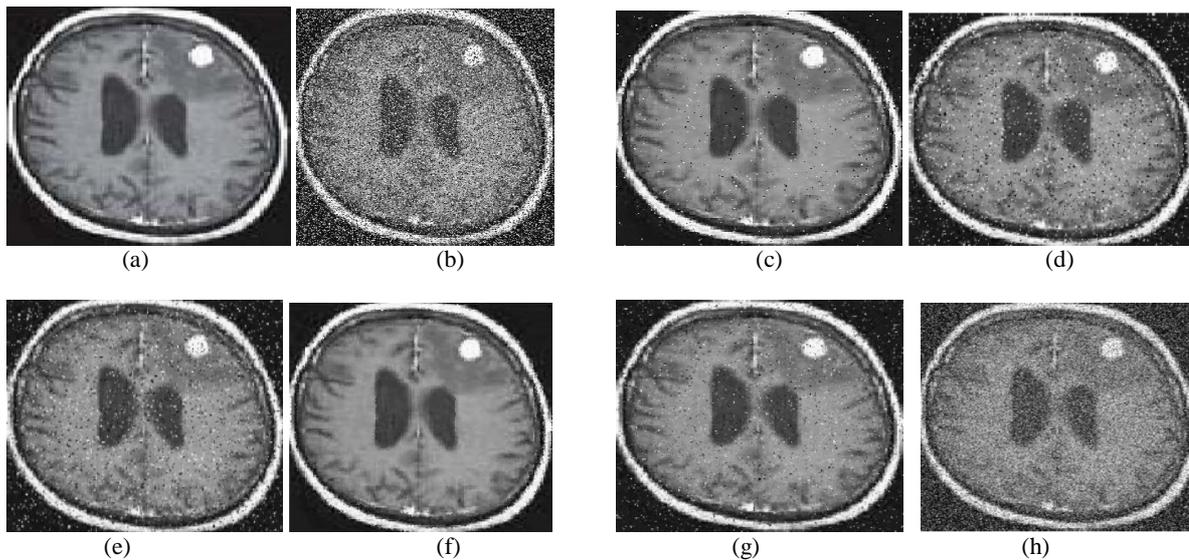

(a)      (b)      (c)      (d)

(e)      (f)      (g)      (h)

*Figure 6* : A set of MRI of brain with different filtering output for the same image Output cited by Manohar Annappa Koli (2012) [20] a) original MRI image, b) original MRI with 50% noise, c) median filter, d) Min-Max Median Filter, e) Center Weighted Median Filter, f) Adaptive Median Filter, g) Progressive Switching Median Filter, h) Average Filter.

**5.2.3 Low pass Filter:** A low pass filter passage[23,24,25] only low value i.e. low intensity value of the pixel.Compute the 2D discrete Fourier transform of the image then apply an image operator to the transformed image and computer the inverse 2D discrete Fourier transform of the result of the image operator. Low-pass filtering zeroes out the frequency components above intensity, if $f(x,y)$ is a function, for example the brightness in an image, its Fourier transform is given by threshold. It suppresses all frequencies higher than the cut-off frequency and leaves smaller frequencies unchanged. This is written below in equation (3).

$$H(u,v) = \begin{cases} D(u,v) & D(u,v) \leq D_0 \\ 0 & D(u,v) > D_0 \end{cases} \qquad D(u,v) = [(u-M/2)^2 + (v-N/2)^2]^{1/2}$$

Take care of narrow noisy variations from MR images.

**5.2.4 High-pass Filter [24, 25, 26]:** working activities of this filter is some other than it passes high frequency and remain unchanged and block low frequency signal, the elements of the mask contain both positive and negative weights.

$$H(u,v) = \begin{cases} 0 & D(u,v) \leq D_0 \\ D(u,v) & D(u,v) > D_0 \end{cases} \qquad D(u,v) = [(u - M/2)^2 + (v - N/2)^2]^{1/2}$$

That is high-pass filter = original – low-pass. High-pass filter take care of undesirable small noise intensity and Useful for emphasizing transitions in intensity (e.g., edges) and Compute intensity differences in local image regions.

**5.2.5 Gaussian Filter:** The Gaussian smoothing [24,25,26] operator for the two dimensional image convolution operators that is used to `blur' images and remove detail and noise. It is produce output similar to the mean filter but some difference are written below. The Gaussian distribution in 2-D has the form is written below in equation (5), where $\sigma$ is the standard deviation of the distribution.

$$G(x,y) = \frac{1}{2\pi\sigma^2} e^{-\frac{x^2+y^2}{2\sigma^2}}$$

Gaussian is random occurrence of white intensity value and its intensity value is drawn from Gaussian distribution, thus it is very much use to reduce Gaussian noise and as it linear filter so it is computationally efficient and enhances image quality with the image boundaries.

**6. Colour Fundamentals:** Colour models or colour spaces, indicate the colours in a benchmark way by using a coordinate system and a subspace in which each colour is represented by a single point of the coordinate system. The largely common colour spaces used in image processing [24, 25] methods are Gray, Binary form, RGB, HSV, HIS etc. Gray form are used in all binarized method, i.e. most of the segmentation technique used binarization methods original MR Brain image is a gray-level image may be inadequate to maintain fine description then pseudo colour conversion are need. Each gray value maps to an RGB item. To obtain more useful features and enhance the visual density, some of the method may applies pseudo-colour transformation, a mapping function that maps a gray-level pixel to a colour level pixel by a lookup table in a predefined colour map. An RGB colour map contains R, G, and B values for each item. But most of the segmentation applied on the gray image the RGB to gray conversion is needed. This colour conversion coordinate is shown in figure below.

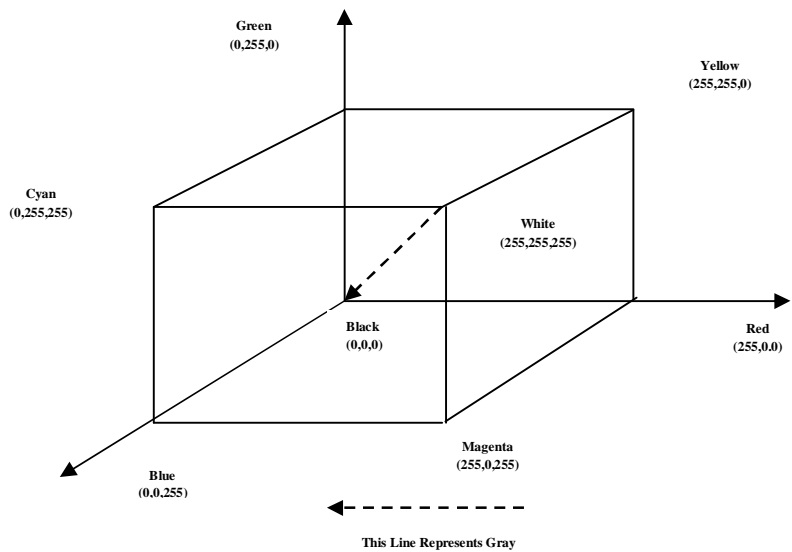

**Figure 7:** Colour fundamental

**7. SEGMENTATION**: Images may be acquired in the continuous domain such as on X-ray film, or in discrete space as in MRI. In 2-D discrete images, the location of each measurement is called a pixel and in 3-D images, it is called a voxel. For simplicity, often it uses the term 'pixel' to refer to both the 2-D and 3-D cases.

If the domain of the image is given by 'I', then the segmentation problem is to determine the sets $S_k \subset I$ whose union is the entire image I. Thus, the sets that make up segmentation must satisfy

$$I = \bigcup_{k=1}^{K} S_k$$

Where $S_k \cap S_j = \emptyset$ for $k \neq j$, and each $S_k$ is connected. Ideally, a segmentation method finds those sets that correspond to distinct anatomical structures or regions of interest in the image.

When the constraint that regions be connected is removed, then determining the sets is called pixel classification and the sets themselves are called classes. Pixel classification rather than classical segmentation is often a desirable goal in medical images, particularly when disconnected regions belonging to the same tissue class need to be identified.

There are several types of segmentation possible to segment a tumor from MRI of brain, those segmentation have several advantages and disadvantages. These advantage and disadvantage have described very carefully with output are describe here. There no such algorithms which always produce very good results for all type of MRI of brain images, thus a brief overview for different type of segmentation are discussed here. Though optimal selection of features, tissues, brain and non–brain elements are considered as main difficulties for brain image segmentation. Thus accurate segmentation over full field of view is another very much problem but during the segmentation procedure verification of results is another source of difficulty. Different segmentation techniques such as thresholding based segmentation methodology, Region Growing based segmentation, K-nearest neighbours (KNN), Bayesian approach, Markov Random Field Models , Expectation maximization (EM), Support vector machine (SVM), Fuzzy c-means algorithms, K-means algorithms, Morphology-based segmentation, Atlas-guided based segmentation, Knowledge based segmentation, Texture-based segmentation, Artificial neural networks (ANNs), Fusion-based, Fuzzy connectedness, Watershed Methods, Level set based segmentation, Hybrid Self Organizing Map (SOM), SOM, Graph Cut based segmentation, Fractal-based segmentation, Parametric deformable models (snakes), Boundary based methods, Geometric deformable model, The Combination of Watershed and Level Set segmentation , Spatio Temporal Model, Hidden Markov Model, Genetic algorithms based segmentation, Kohonen Self Organizing Map(SOM) are describe in this section.

**7.1 Threshold Based Segmentation:** Threshold is one of the aged procedures for image segmentation. These threshold techniques are very much useful for image binarization which is very essential task for any type of segmentation. It assumes that images are composed of regions with different gray level ranges. A thresholding procedure determines an intensity value, called the threshold, which separates the desired classes [31]. There are several threshold segmentation techniques exist, among them here describe some well known and well established thresholding techniques such as Otsu method , Bernsen method, Sauvola method , Niblack method , Kapur method , Th-mean method, Iterative as frame work to all existing method, Balance Histogram which is describe below.

**7.1.1 Otsu method:** This is a global thresholding method .In this method [32], the threshold operation is observed as the partitioning of the pixels of an image into two classes objects and background at gray level *t*. Calculated by the within-class variance, between-class variance, and the total variance, respectively. This algorithms does not work properly for all type MRI of brain image, this is because of large intensity variation of the foreground and background image intensity. Otsu binarize whole image and produces some unnecessary part. Here test all the methods about 30 images out of them  some output of Otsu methods are shown below, thus figure it is clear that Otsu methods is not suitable for brain tumor segmentation and quantification because it binarize whole image and produce an output which is not meaningful.

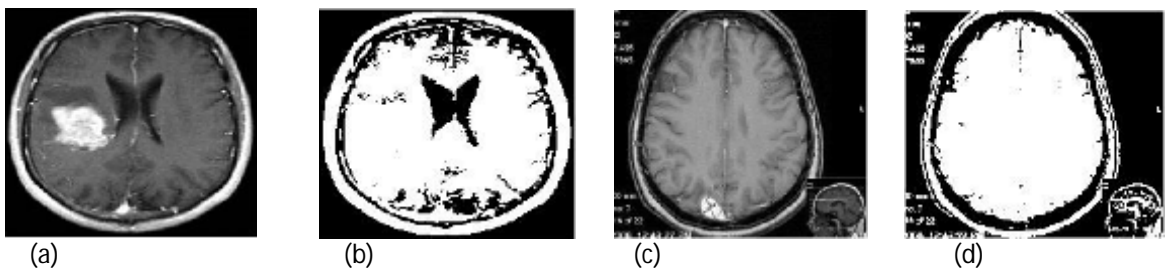

(a) (b) (c) (d)

**Figure 8:** Shows of Otsu binarization methods: a) Original MRI image, b) output of otsu method, c) Original MRI image, d) output of Otsu method.

**7.1.2 Th-mean method:** Th-mean algorithms [37] approach is the determining of thresholding of small region of the image and actual selection of threshold had done by mean of the all the thresholds. From the output produce by Th-mean methods it concludes that this method not a suitable for MRI of brain tumor segmentation.

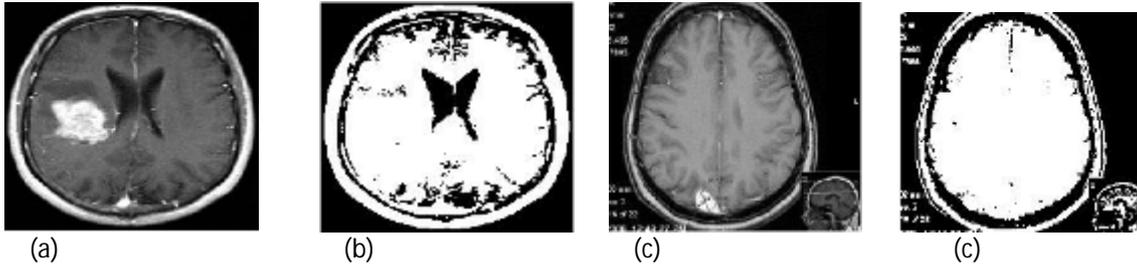

(a)  (b)  (c)  (c)

*Figure 9:* Shows of Th-mean binarization methods: a) Original MRI image, b) output of Th-mean method, c) Original MRI image, d) output of Th-mean method.

**7.2 Region Growing:** It requires a seed point that is manually selected by the user and removes all pixels connected to the preliminary seed based on some predefined condition. It is a procedure for extracting an image region that is connected based on some predefined criterion. These conditions can be based on intensity information or boundaries in the image [42]. The possible criterion might be to grow the region until an boundary in the image is met. Region increasing is seldom used alone but usually within a set of image processing operations, mostly for the description of small, simple structures such as tumors and lesions [42, 43, 44]. The manual dealings to obtain the seed point is the great disadvantage for this region growing. Thus, for each region that needs to be extracted, a seed must be planted but split-and-merge is an algorithm related to region growing, but it does not require a seed point [45, 46]. Region growing has also been restriction to susceptible to noise i.e. very much sensitive to noise; causing extracted regions to have holes or even become disjointed. These problems may overcome by using a hemitropic region-growing algorithm [47, 48]. The region growing method is a well-developed technique for image segmentation. The technique is not fully automatic [49, 50], i.e. it requires user interaction for the selection of a seed and secondly the method fails in producing acceptable results in a natural image i.e. Only works in homogeneous areas. Since this technique is noise sensitive, therefore, the extracted regions might have holes or even some discontinuities.

**7.3 K-nearest neighbours (KNN):** The KNN classifier is considered a non-parametric classifier since it makes no underlying assumption about the statistical structure of the data [51]. The k-NNR only requires an integer k, set of labelled examples (training data) and a metric to measure closeness by Euclidean distance

$$d(Xu, X) = \sqrt{(\sum_{k=1}^{D} (Xu(k) - X(k))^2)}$$

For example there are three classes and the goal is to find a class label for the unknown example $X_u$ and in this case we use the Euclidean distance and a value of k=5 neighbours i.e. 5 closest neighbours, thus from the figure below it has seen that 4 belong to $\omega_1$ and 1 belongs to $\omega_3$, so $X_u$ is assigned to $\omega_1$.

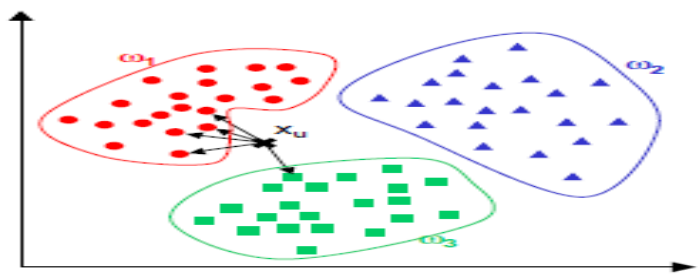

*Figure 10*: A conceptual diagram with k=5 neighbour, $w_1$, $w_2$, $w_3$ are three different region and Xu be the point from where we select 5 point. 4 point are selected from $w_1$ and 1 point is selected from $w_2$.

k-NN is very simple to understand and easy to implement [51, 52]. So it should be considered in seeking a solution to any classification problem. Some advantages of k-NN are it is easy to implement and debug, in situations where an explanation of the output of the classifier is useful, k-NN can be very effective if an analysis of the neighbours is useful as explanation and there are some noise reduction techniques that work only for k-NN that can be effective in improving the accuracy of the classifier. Disadvantages are k-NN can have poor run-time performance if the training set is large because all the work is done at run-time, k-NN is very sensitive to irrelevant or redundant features because all features contribute to the similarity and thus to the classification and this can be ameliorated by careful feature selection or feature weighting [53,54]. On very difficult classification tasks, k-NN may be outperformed by more exotic techniques such as Support Vector Machines or Neural Networks. A simple classifier is the KNN classifier, where each pixel or voxel is classified in the same class as the training data with the closest intensity. This method did not include any spatial regularization, so it is very sensitive to noise and in homogeneity of tumors. The KNN classification and registration of anatomical brain atlas are then iterated to improve the result of classification [55, 56].

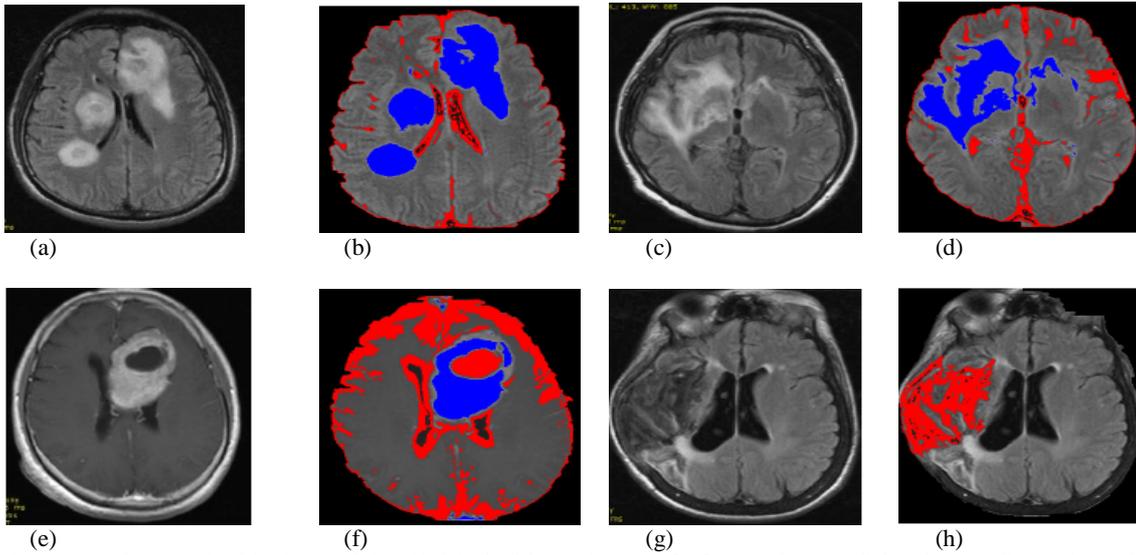

*Figure 11*: Output cited by Noor Elaiza Abdul Khalid et. al.(2011) [52] a) and c) are light abnormalities , b) and d) are there output using k-NN similarly e) and g) are dark abnormalities , f) and h) are there output using k-NN, thus k-NN on the other hand performed well in dark abnormalities segmentation

This system provides good results for small tumors but in the case of large deformations in the brain it will fail. This method also needs much calculation by repeating the classification and registration, therefore it is relatively slow. This algorithm fails in cases where the intensity distribution in the tumor is highly inhomogeneous and shows large spectral overlap with brain tissues. Other disadvantage of this KNN algorithm include the dependence on the parameter K, large storage requirements (for training points), sensitivity to noise in the training data, and the undesirable behaviour that can occur in cases where a class is under-represented in the training data, which make it unsuitable for brain tumor segmentation in MRI.

**7.4 Bayesian approach:** This is a supervised and parametric approach, where the data are assumed to follow a multivariate normal distribution, where mean and covariance are estimated from the training data set [57]. This method combines a graph-based algorithm and Bayesian model and segments the edema in addition. Also it can be extended to vectorial variables to operate on multi-modality images.

A Bayesian network is a model of compound probability distribution function of a set of variable like directed acyclic graph with a probability table for each node. The nodes in a Bayesian network depends upon different variables in a domain, and the arcs between nodes represent the dependency relationships among the variables with probability [57]. First introduce some terms are very shortly written. Prior probability is the probability determined by the historic materials or the judgment of somebody. Since this kind of probability is not validated by experiment, it is called prior probability. Posterior probability is the probability which revised according to Bayesian equation and new features achieved by analysis thus the total probability theorem if A can be only influenced by factors $B_i$. $B_j = \emptyset$ (i≠j), $P(B_i) > 0$, i = 1, 2,...then it have:

$$P(A) = \sum P(B_i)P(A/B_i)$$

Bayesian Equation is also called posterior probability equation because in this probability distribution some other investigations are needs. If the prior probability is $P(B_i)$ and the additional information obtained by investigation is $P(A|B_i)$, where i = 1,2,..., n, then the probability is:

$$P\left(B_i/A\right) = \frac{P(B_i)P(A/B_i)}{\sum_{k=1}^{n} P(B_k)P(A/B_k)}$$

The Bayesian Networks B = {G, Θ} is a directed graph with a probability to each node with i)each node of the network represents a variable which can be discrete or continuous, ii)A set of directed edges or arrows, if there exists an arrow from node X to node Y, then X is called the parents node of Y, iii) For each node $X_i$, there is a conditional probability distribution $P(X_i/Pa_i)$, which indicate the influence by its parents, the graph must be directed acyclic graph. Then a Bayesian network defines a joint probability distribution written as follows

$$P_B(X_1,\ldots,X_n) = \prod_{i=1}^{n} P_B(X_i/Pa_i)$$

The goal is to estimate the class labels by maximizing the a posteriori probability by observed data and its class. The other problem of this method is the modelling of the tumor by a Gaussian model, since the probability of tumor and edema do not always follow Gaussian distributions.

The prior probabilities for the normal tissue classes white matter, gray mater and other are defined by the registered spatial atlas to the patient images and the tumor spatial prior is calculated from the difference image are converted to probability values through histogram analysis[58]. This method segments only the full-enhanced tumors and in the case of the presence a large deformations in the brain it fails. In addition the probability distribution of tumor and edema has been assumed to be a normal distribution and it is not correct in the all cases [59]. In the case of edema, the authors have assumed a fraction of white matter probability for edema, although we cannot always consider this fraction. Jason J. Corso (2008) et. al. [60]integrate the resulting model-aware affinities into the multilevel segmentation by weighted aggregation algorithm, and apply the technique to the task of detecting and segmenting brain tumor and edema in multichannel MR volumes. The computationally efficient method runs orders of magnitude faster than current state-of the- art techniques giving comparable or improved results. In most cases Jason J. Corso (2008) et. al. [60] show a single, indicative slice from the volume, all processing is in three dimensions and the results show good segmentation and classification on a comparatively large dataset with 70% accuracy.

**7.5 Markov Random Field Models**: Markov random field theory holds the promise of providing a systematic approach to the analysis of images in the framework of Bayesian probability theory. Markov random fields (MRFs) model the statistical properties of images. This allows a host of statistical tools and approaches to be turned to solving so called ill-posed problems in which the measured data does not specify a unique solution [61,62]. Markov Random Field Models (MRFs) are not deterministic, are best characterized by their statistical properties. For example, textures can be represented by their first and second statistics. Images are often distorted by statistical noise. To restore the true image, images are often treated as realizations of a random process. They can be used to model spatial constrains such as smoothness of image regions, spatial regularity of textures in a small region and depth continuity in stereo construction. MRFs [63] is depends upon basically two basic concepts, Neighbors and cliques. Let S be a set of locations, here for simplicity, assume S a grid. S={ (i, j) | i, j are integers }. Neighbors of $s(i,j) \in S$ are defined as: $\partial((i, j)) = \{ (k, l) \mid 0<(k - i)^2 + (l - j)^2 < r \text{ constant} \}$, r is constant . A subset C of S is a *clique* if any two different elements of C are neighbors. The set of all cliques of S is denoted by Ω. Thus a four neighbor with radius one, graph and cliques are shown below.

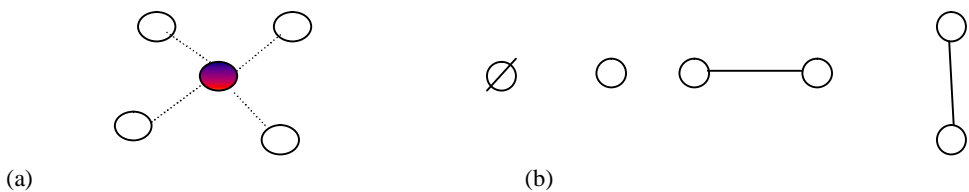

(a)   (b)

*Figure 12:* a) shows 4-neighborhood with radius equals to 1 and b) shows there cliques.

The random vector $X = \{X_s\}_{s \in S}$ on S is called a random field and assumed to have density p(x). Images as Random fields: If vector X represents intensity values of an image, then its component $X_s$ is the intensity value at location s=(i, j). If p(x) of a random field fulfills the so called Markov condition with respect to a neighborhood system, it is called a Markov Random Field.

$$p(X_s = x_s \mid X_t = x_t, \forall t \neq s) = $$
$$p(X_s = x_s \mid X_t = x_t, \forall t \in \partial\{s\})$$

The value $X_s$ at location S is only depend on its neighbors. p(x) can also be factorize over cliques due to its Markov properties. i.e.

$$p(x) = \prod_{C \in \Omega} \Psi_c(x) \quad I \subset S$$

$\Psi_C$ is a function of X determined by clique C. For every , then

$$P(X_I = y_I \mid X_{S \setminus I} = x_{S \setminus I}) =$$
$$Z_I^{-1} \exp(-H(y_I x_{S \setminus I}))$$

S\I means complement of I. If I is a small set, since X only changes over I, $Z_I$ can be evaluated in reasonable time.
So $p(y_I|x_{S\setminus I})$ is known[63]. Using MRFs in Image Analysis, In image analysis, p(x) is often the posterior probability of Bayesian inference which is discuss above i.e. p(x) = p(x|y_0). For example, $y_0$ may be the observed image with noise, and to compute the estimate $x_0^*$ of the true image $x_0$ based on p(x) = p(x|y_0). Thus the difficulties in computing $X_0^*$ is a way to compute the estimate $X_0^*$ is to let

$$x_0^* = E(x)_{p(x)} = \int x p(x \mid y_0) dx$$

But p(x|y_0) is only known up to a constant Z, thus the above integration too much difficult and it can be slow if there are many rejections..
One version of above called Gibbs Sampler[63], builds the Markov chain and updates only a single component of $X^t$ in one iteration. Let the vector X has k components, $X=(X_0,X_1,X_2,......,X_k)$ and presently it is in state $X^t = (x_0,x_1, x_2,......,x_k)$.
An index that is equally likely to be any of 1,......,k is chosen. say index i. A random variable w with density P{w=x} = P{$X_i$=x | $X_j$ = $x_j$, j ≠ i } is generated. If w=x, the updated $X^t$ is $X^{t+1}$ = $(x_0, x_1, x_2, ..., x_{i-1}, X, x_{i+1}, ..., x_k)$.
These local correlations provide a mechanism for modelling a variety of image properties. In medical imaging, they are typically used to take into account the fact that most pixels belong to the same class as their neighbouring pixels. MRF is a statistical model which can be used within segmentation methods [64]. This is only applicable to tumors that are homogeneous enough to be segmented into a single normal tissue class, therefore is not generally applicable to heterogeneous tumors and it allows the identification of tumor structures that have normal intensities but are too thick to be normal.

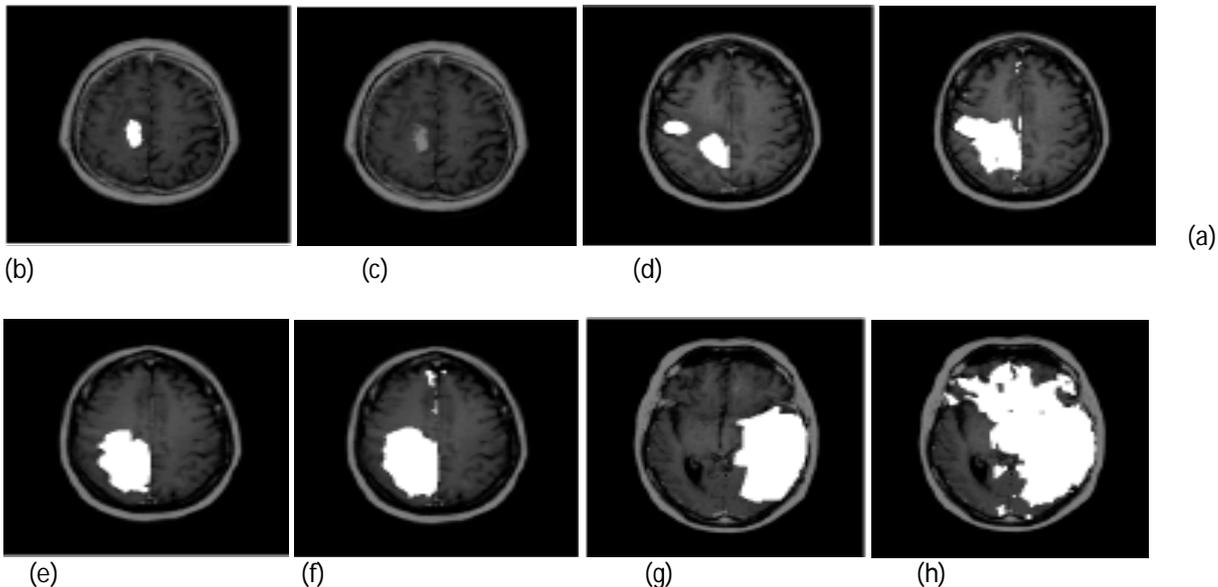

(b) (c) (d) (a)

(e) (f) (g) (h)

*Figure 13*: Output cited by Chi-Hoon Lee et. al [64]. a), c), e) and g) are the tumor segmented by expertise radiologist, and b), d), f), g) are the corresponding brain tumor/ edma ground truth by the Markov random field methods.

**7.6 Support Vector Machine (SVM):** The SVM approach is considered as a good candidate due to high generalization performance, especially when the dimension of the feature space is very high. The SVM uses the following idea: it maps the input vector x into a high-dimensional feature space Z through some non-linear mapping, chosen a priori [67]. SVM became famous when, using images as input, it gave accuracy comparable to neural-network with hand-designed features in a handwriting recognition task and currently, SVM is widely used in object detection & recognition, content-based image retrieval, text recognition, biometrics, speech recognition, etc[64]. Those training points for which the equality in of the separating plan is satisfied (i.e. $y_i (x_i \cdot w + b) -1 \geq 0, \forall i$), those which wind up lying on one of the hyper planes $H_1$, $H_2$), and whose removal would change the solution found, are called *Support Vectors (SVs)* and are indicated in Figure 18 by the extra circles above the dotted line.

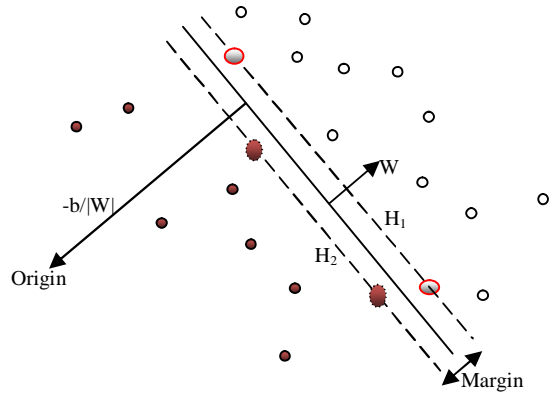

*Figure 14*: *Linear separation of hyper planes for the separable cases and the support vector are circled in the dotted line.*

The classifier is said to assign a feature vector *x* to class $w_i$ if

$$g_i(\mathbf{x}) > g_j(\mathbf{x}) \quad \text{for all} \quad j \neq i$$

For two-category case,

$$g(\mathbf{x}) \equiv g_1(\mathbf{x}) - g_2(\mathbf{x}) \quad \text{Decide } \omega_1 \text{ if } g(\mathbf{x}) > 0; \text{ otherwise decide } \omega_2$$

Thus the minimum error-rate classifier

$$g(\mathbf{x}) \equiv p(\omega_1 | \mathbf{x}) - p(\omega_2 | \mathbf{x})$$

Linear Discriminant Function g(x) is a linear function is given by

$$g(\mathbf{x}) = \mathbf{w}^T \mathbf{x} + b$$

A hyper-plane in the feature space unit-length normal vector of the hyper-plane $\quad \mathbf{n} = \dfrac{\mathbf{w}}{\|\mathbf{w}\|}$

Nonlinear SVMs: The Kernel Trick with this mapping, our discriminant function is

$$g(\mathbf{x}) = \mathbf{w}^T \phi(\mathbf{x}) + b = \sum_{i \in SV} \alpha_i \phi(\mathbf{x}_i)^T \phi(\mathbf{x}) + b$$

Thus no need to know this mapping explicitly, because it only use the dot product of feature vectors in both the training and test. A kernel function is defined as a function that corresponds to a dot product of two feature vectors in some expanded feature space is written below

$$K(\mathbf{x}_i, \mathbf{x}_j) \equiv \phi(\mathbf{x}_i)^T \phi(\mathbf{x}_j)$$

Support Vector Machines (SVMs) [68] are a popular tool for classification of data that is independent and identically distributed. SVMs try to maximize the margin between classes i.e. here using the simple linear feature space xi ·xj , by finding the optimal values in the following Quadratic Programming problem represented in dual Lagrangian form where C is a constant that bounds the misclassification error

$$\text{maximize} \sum_{i=1}^{n} \alpha_i - \frac{1}{2} \sum_{i=1}^{n} \sum_{j=1}^{n} \alpha_i \alpha_j y_i y_j K(\mathbf{x}_i, \mathbf{x}_j)$$

Where

$$0 \leq \alpha_i \leq C \qquad \sum_{i=1}^{n} \alpha_i y_i = 0$$

Unlabelled instances are classified using the learned parameters $\alpha_i$ and bias b, by taking the sign of the following decision function

$$g(\mathbf{x}) = \sum_{i \in SV} \alpha_i K(\mathbf{x}_i, \mathbf{x}) + b$$

Here we give some examples of commonly-used kernel functions like linear, polinomial, Gaussian kernel etc. Thus to obtained a Support Vector Machine segmentation algorithm we must i) choose a kernel function, ii) Choose a value for C, iii) Solve the quadratic programming problem (many software packages available, iv) Construct the discriminant function from the support vectors [69]. Choice of kernel may have Gaussian or polynomial kernel is default but if ineffective, more elaborate kernels are needed then domain experts can give assistance in formulating appropriate similarity measures. σ in Gaussian kernel may choose Gaussian kernel parameter where σ is the distance between closest points with different classifications. In the absence of reliable criteria, applications rely on the use of a validation set or cross-validation to set such parameters.To optimization a lengthy series of experiments in which various parameters are tested. Different of SVM are shown below.

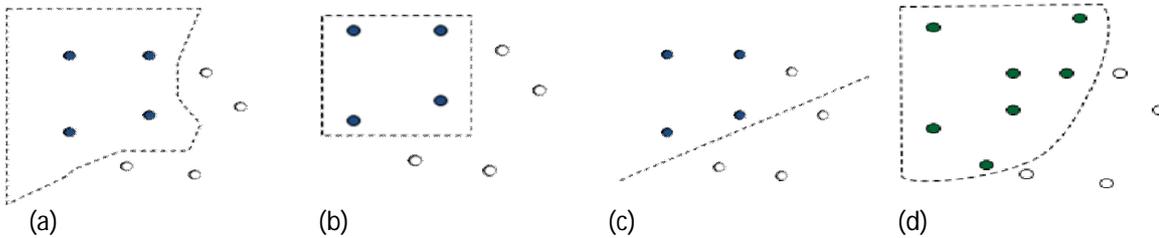

(a)    (b)    (c)    (d)

*Figure 15: Shows the different type of SVM a) Nearest Neighbor, b) Decision Tree, c) Linear Functions, d) Nonlinear Functions has been shown dotted line.*

A SVM classification to classify the brain into the tumor and non-tumor classes using T1-weighted and contrast enhanced T1-weighted images. Some morphological operations have been used to eliminate the classification inaccuracy. This system used patient-specific training and compared two different types of SVM, the standard 2-class method and the more recent 1-class method [70,71]. The advantage of using a one-class method has elimination in the manual time needed to carry out patient specific training, since only training examples for the tumor class were needed.

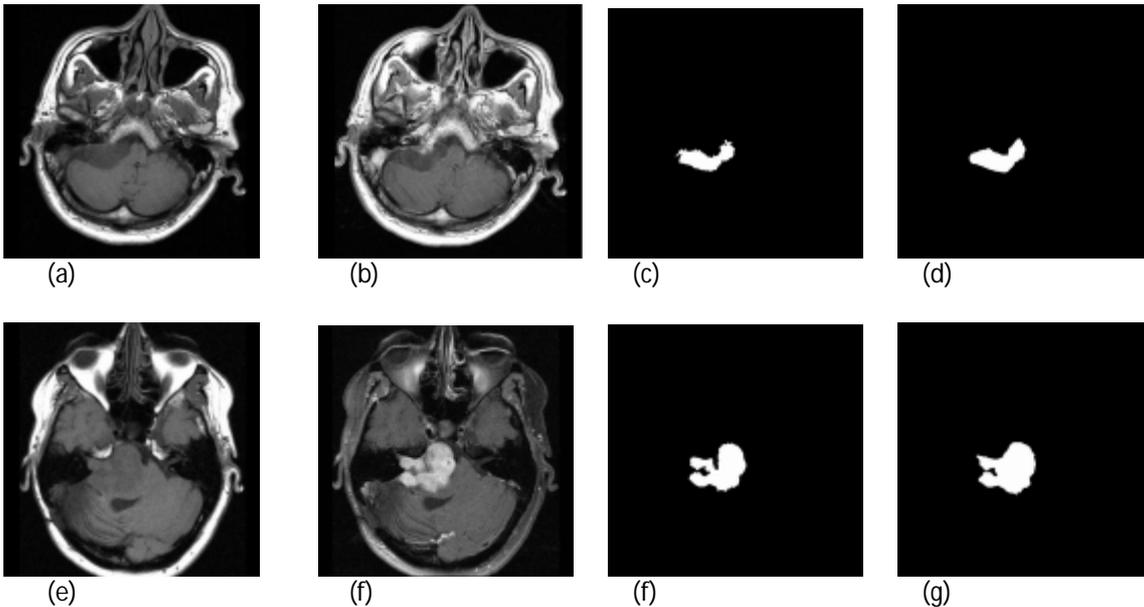

(a)    (b)    (c)    (d)

(e)    (f)    (f)    (g)

*Figure 16: Output cited by J. Zhou et.al. (2005)[70]. examples of brain tumor segmentation results from two patients and the corresponding ground truths. a) and e) original T1W brain MR images before, b) and f) after contrast enhancement, c) and f) segmentation results using SVM one-class method, d) ground truth by expertise radiologist.*

The SVM method has the advantage of generalization and working in high dimensional feature space, it assumes that data are independently and identically distributed which is not appropriate for tasks such as segmenting medical images with inhomogeneity and noise and so it must be combined with other methods to consider spatial information and also have the advantages of such classifiers are that they are independent of the dimensionality of the feature space and that the results obtained are very accurate, although the training time is very high. In addition the problem of patient-specific learning and storage must be added to the disadvantage of SVM-based methods. We also see that regular one class SVMs do not consider the negative information which cannot learn the feedback well.

**7.7 Fuzzy c-means algorithms:** The goal of a clustering analysis is to divide a given set of data or objects into a cluster, which represents subsets or a group. The partition should have two properties one of them is the homogeneity inside clusters data, which belongs to one cluster, should be as similar as possible and another one is heterogeneity between the clusters data, which belongs to different clusters, should be as different as possible [73]. The membership functions do not replicate the actual data distribution in the input and the output. They may not be suitable for fuzzy pattern recognition. To build membership functions from the data available, a clustering technique may be used to partition the data, and then produce membership functions from the resulting clustering [74]. The FCM algorithm is an improvement of earlier clustering methods. The objective function of FCM algorithm is defined as the sum of distances between the patterns and the cluster centers.

$$J = \sum_{i=1}^{M}\sum_{j=1}^{C} U_{ij}^{q} d(x_i, v_j)$$

Where, C is the number of clusters that in this problem is equal to number of needed segmentation regions. The represents the center of cluster j; M is the number of brain MR image pixels; the parameter q is larger than 1, and adjusts the fuzzifier intensity; is the membership function of attribute to cluster j, which should satisfy these conditions:

$$U_{ij} \in [0,1]; \quad \sum_{j=1}^{c} U_{ij} = 1; \quad 0 < \sum_{i=1}^{M} U_{ij} < M$$

The $d(x_i, v_j)$ measures the dissimilarity between $x_i$ and $v_j$. The popular selection for $d(x_i, v_j)$ is

$$d(x_i, v_j) = \|x_i - v_j\|^2 = \|x_i - v_j\|^2 * w_f * \|x_i - v_j\|^2$$

Here, $W_f$ is a positive symmetric definite matrix. In order to optimize, the gradient descent can be used.

$$U_{ij} = 1 / \sum_{k=1}^{c} \left( \frac{d(x_i, v_j)}{d(x_i, v_k)} \right)^{(2/(q-1))}$$

And the cluster center $v_j$ is defined as:

$$v_j = \left( \sum_{i=1}^{M} U_{ij}^q x_i \right) / \left( \sum_{i=1}^{M} U_{ij}^q \right)$$

Note that brain MR images are gray scale images, so the dimension of feature vectors is equal to one. The FCM algorithm can optimize the objective function J using in an interactive way. To terminate this process, an ending criterion such as U(t) – U(t-1) < € can be used.

The traditional FCM clustering algorithm for MR images segmentation, which may performs very fast and simple, but this algorithm do not guarantee high accuracy especially for noisy or abnormal images. Unfortunately,[75] MR images always contain a significant amount of noise caused by operator, equipment, and the environment, which lead to serious inaccuracies in the segmentation[76]. It should be mentioned that the membership functions to classes have a counter intuitive shape, which limits their use, fuzzy C means is a popular method for medical image segmentation but it only considers image intensity thereby producing unsatisfactory results in noisy images.

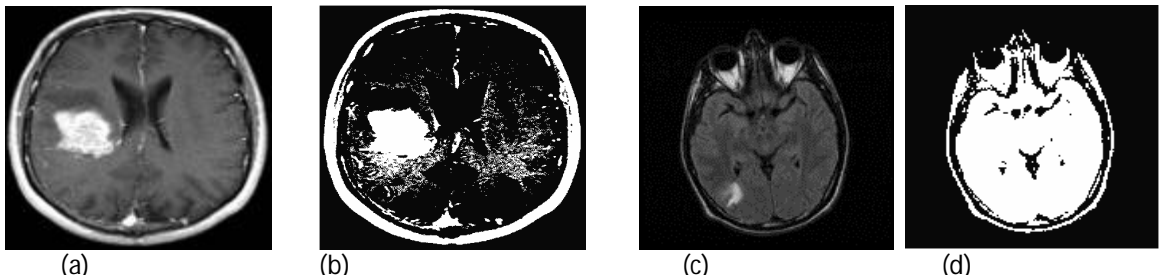

(a)　　　　　(b)　　　　　(c)　　　　　(d)

*Figure 17:* Brain tumor segmentation by fuzzy C-means results a) and c) original brain MR images, b) and d) output produce by fuzzy c-means: For the first image tumor detect but produce noise and for second image abnormality are not detected.

A bunch of algorithms are proposed to make FCM robust against noise and in homogeneity but it's still not perfect [77, 78, 79]. To solve this problem, Shen et al., 2003 [80] have proposed a more recent system, which incorporated intensity standardization (using the pixel histograms) as a pre-processing step, and a modified FCM algorithm which involves dependencies between neighbour pixels. This method is more robust to noise and provides a better segmentation quality in comparison with the other FCM based approaches.

**7.8 K-means algorithms:** K-Means clustering [84] generates a specific number of disjoint, flat (non-hierarchical) clusters. It is well suited to generating globular clusters. The K-Means method is numerical, unsupervised, non-deterministic and iterative. K-Means Algorithm are i) always K clusters, ii) always at least one item in each cluster, iii)The clusters are non-hierarchical and they do not overlap, iv) Every member of a cluster is closer to its cluster than any other cluster because closeness does not always involve the center of clusters. Thus the K-Means Algorithm Process in very shortly a)The dataset is partitioned into K clusters and the data points are randomly assigned to the clusters resulting in clusters that have roughly the same number of data points, b) For each data point: Calculate the distance (Mahalanobis or Euclidean) from the data point to each cluster, c) If the data point is closest to its own cluster, leave it where it is and if the data point is not closest to its own cluster, move it into the closest cluster, d) Repeat the above step until a complete pass through all the data points results in no data point moving from one cluster to another. At this point the clusters are stable and the clustering process ends, e)The choice of initial partition can greatly affect the final clusters that result, in terms of inter-cluster and intra-cluster distances and cohesion. The algorithm [84, 85, 86] assumes that the data features form a vector space and tries to find natural clustering in them. The points are clustered around centroids $\mu_i \forall_i = 1 \ldots \ldots k$ which are obtained by minimizing the objective.

$$V = \sum_{i=1}^{k} \sum_{x_j \in S_i} (x_i - \mu_i)^2$$

Where there are k clusters $S_i$, i = 1,2, ......, k and $\mu_i$ is the centroid or mean point of all the points $x_j \in S_i$. As a part of this project, an iterative version of the algorithm was implemented. The algorithm takes a 2 dimensional image as input. It computes the intensity distribution then initializes the centroids with k random intensities. Repeat until the cluster labels of the image does not change anymore [84]. Cluster the points based on distance of their intensities from the centroid intensities.

$$c^{(i)} = \arg\min over\ j\ |x^{(i)} - \mu_j|^2$$

Then compute the new centroid for each of the clusters.

$$\mu_i = \frac{\sum_{i=1}^{m} 1\{c^{(i)} = j\} x^{(i)}}{\sum_{i=1}^{m} 1\{c^{(i)} = j\}}$$

Where k is a parameter of the algorithm denoted by the number of clusters to be found in the set and $\mu_i$ are the centroid intensities. The formula used to calculate Mahalanobis distance [87,88, 89] is : Dt(x) = (x – $C^i$) * Inverse(S) * (x – $C^i$). Here X is a data point in the 3-D RGB space, $C^i$ is the center of a cluster, S is the covariance matrix of the data points in the 3-D RGB space, Inverse(S) is the inverse of covariance matrix S. The formula used to calculate Euclidean Distance: The Euclidean distance is the straight-line distance between two pixels = √(($x_{i1} - x_{j2}$)² + ($y_{i1} - y_{j2}$)²) , where ($x_{i1},y_{j1}$) and ($x_{j2},y_{j2}$) are two pixel points or two data points.

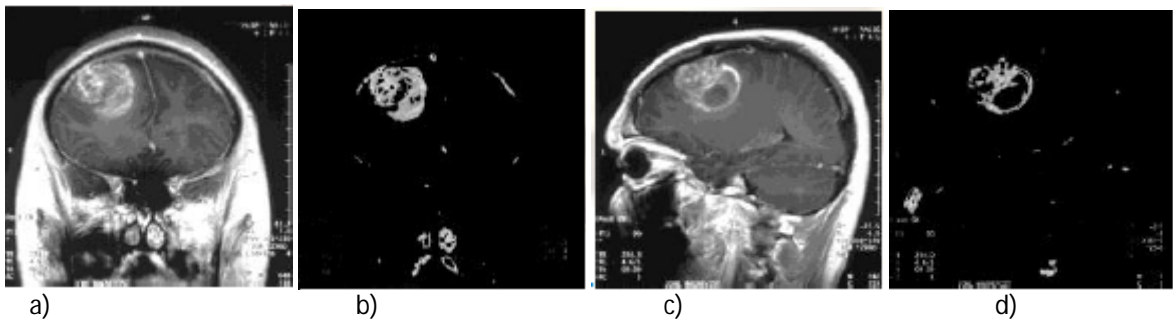

a)  b)  c)  d)
*Figure 18* : a) and c) are the input MRI image; and b) and d) are corresponding K-means output

K-means fairly simple to implement and image segmentation are impressive [90,91]. As can be seen by the results, the number of partitions used in the segmentation has a very large effect on the output [93]. By using more partitions, in the RGB setup, more possible colors are available [93] in the output. The same is true for the setup checking greyscale intensity. By adding more partitions, a greater number of intensities are available to use in the output image. The algorithm also runs quickly enough that real-time image segmentation could be done with the K-Means algorithm.

The k-means clustering algorithm [94, 95] clusters data by iteratively computing a mean intensity for each class and segmenting the image by classifying each pixel/voxel in the class with the closest mean consists of unsupervised classification of patterns into groups (clusters).The clustering algorithms essentially work such as classification methods without use of training data set.

**7.9 Morphology-based:** Morphological image processing [96] (or morphology) describes a range of image processing techniques that deal with the shape (or morphology) of features in an image and morphological operations are typically applied to remove imperfections introduced during segmentation, and so typically operate on bi-level images i.e. binary images [97]. Using some simple technique it has looked at so far we can begin to consider some more interesting morphological algorithms. It use morphological operation in boundary extraction, Region filling, extraction of connected components, thinning/thickening, skeletonisation [98]. All morphological processing operations are based on these simple ideas like *Fit*: All on pixels in the structuring element cover on pixels in the image and *Hit*: Any on pixel in the structuring element covers an on pixel in the image with structuring element [99]. Structuring elements can be any size and make any shape. However, for simplicity it uses rectangular structuring elements with their origin at the middle pixel. Fundamentally morphological image processing is very like spatial filtering and the structuring element is moved across every pixel in the original image to give a pixel in a new processed image and the value of this new pixel depends on the operation performed[100]. There are two basic morphological operations: erosion and dilation [100, 101,102,103]. Erosion of image f by structuring element s is given by $f \ominus s$. The structuring element s is positioned with its origin at (x, y) and the new pixel value is determined using the rule:

$$g(x, y) = \begin{cases} 1 & \text{if } s \text{ fits } f \\ 0 & \text{otherwise} \end{cases}$$

Dilation of image f by structuring element s is given by $f \oplus s$ The structuring element s is positioned with its origin at (x, y) and the new pixel value is determined using the rule:

$$g(x, y) = \begin{cases} 1 & \text{if } s \text{ hits } f \\ 0 & \text{otherwise} \end{cases}$$

More interesting morphological operations can be performed by performing combinations of erosions and dilations thus the basic concept is to search an image with a structuring element and to quantify the manner in which the structuring element fits within the image [104]. Region growing can also be sensitive to noise, causing extracted regions to have holes or even become disconnected. Also the major problems of the region growing are the leakage of the segmented volume into adjacent structures because of the weak border of tumor [105, 106]. Segment only the solid section of the tumor and segmentation of the other components of tumor such as edema and necrosis may not considered.

**7.10 Atlas-guided based:** This atlas is created by manual segmentation or by other semi-automatic segmentation methods. Atlas can capture spatial, intensity and shape distributions of the anatomical structures of interest [107]. This atlas is then used as a reference frame for segmenting new images. Atlas-guided approaches have been applied mainly in MR brain imaging. An advantage of atlas-guided approaches is that labels are transferred as well as the segmentation. They also provide a standard system for studying morphometric properties[107,108,109]. Even with non-linear registration methods however, accurate segmentations of complex structures is difficult due to anatomical variability. The atlas-based segmentation has an ability to segment the image with no well defined relation between regions and pixels' intensities[110]. This can be due to lack of the border or excessive noise or in the case when the objects of the same texture need to be segmented. If the information about difference between these object is incorporated in spatial relationship between them, other objects, or within their morphometric characteristics, the atlas-based segmentation is expected to work well. Another important advantage of atlases is in their use in clinical practice, for computer aided diagnosis whereas they are often used to measure the shape of an object or detect morphological differences between patient groups.
On the other hand the disadvantage of an atlas-based can be in the time necessary for atlas construction wherever iterative procedure is incorporated in it, or a complex nonrigid registration [111]. Since the atlas based segmentation is usually used when the information from the gray level intensities are not sufficient, it is difficult to produce objective validation. With proper selection of an atlas and enough plastic transformation the different segmentation properties may be achieved. If use the atlas based segmentation for segmentation of an object within image, then define the registration procedure.
This atlas is created by manual segmentation or by other semi-automatic segmentation methods [112]. Atlas can capture spatial, intensity and shape distributions of the anatomical structures of interest. This atlas is then used as a reference frame for segmenting new images. A global transformation or registration technique is used to align the atlas to the new image that will be segmented and then the atlas information will be applied to refine the segmentation or to detect abnormalities in the image [113, 114]. Therefore these types of segmentation deal also with registration problems and the quality of segmentation depends on the registration method. The standard atlas-guided approach treats segmentation as a registration problem. It first finds a one-to-one transformation that maps a pre-segmented atlas image to the target image that requires segmenting. This process is often referred to as atlas warping. A few exceptions Bach et al. [115]) these studies have been conducted on image volumes that do not contain tumors or lesions which can alter the anatomy of the brain considerably. The objective of the study presented herein is to assess the validity of atlas-based techniques for the automatic segmentation of most structures delineated for typical radiation therapy applications.

**7.11 Knowledge based:** The system evolves from clustering the entire image to clustering very specific areas, while the rules are used to remove the clusters that do not have tumor properties. Fletcher-Heath et al., 2001[116] have later developed this algorithm for segmentation of non-enhanced brain tumors in MRI by changing the rules of the expert system. Drawback of this type is that the rules may not be robust to nonstandard intensity and the errors can propagate if the assumptions of early rules in the sequence are violated. Another disadvantage of this approach is the considerable manual engineering requirement. This is due to the difficulty of translating complex anatomic knowledge and visual analysis into sequential low-level operations and rules. Also in the case of heterogeneous tumor or noise, the rules may not work correctly[117]. The value of each voxel intensity in a given class can be considered as a random variable, independent across pixels. In the following results, we assume that the voxel intensities are normally distributed. This assumption may be modified to support other distributions that may better fit the data. With a large set of training data, the distributions may also be learned a priori. John Melonakos et al.[118] proposed a user specifies number of classes i.e. 'N' (default $N = 2$) and find N initial class means and standard deviations using K-Means clustering then Generate N images of prior terms after this generate N images of data terms and then Apply Bayes' Rule to prior and data images to obtain N posterior images then smooth the posterior images for several iterations using an edge-preserving PDE and renormalize the posterior images after each smoothing iteration then Apply maximum a posteriori rule to achieve segmentation labeling.

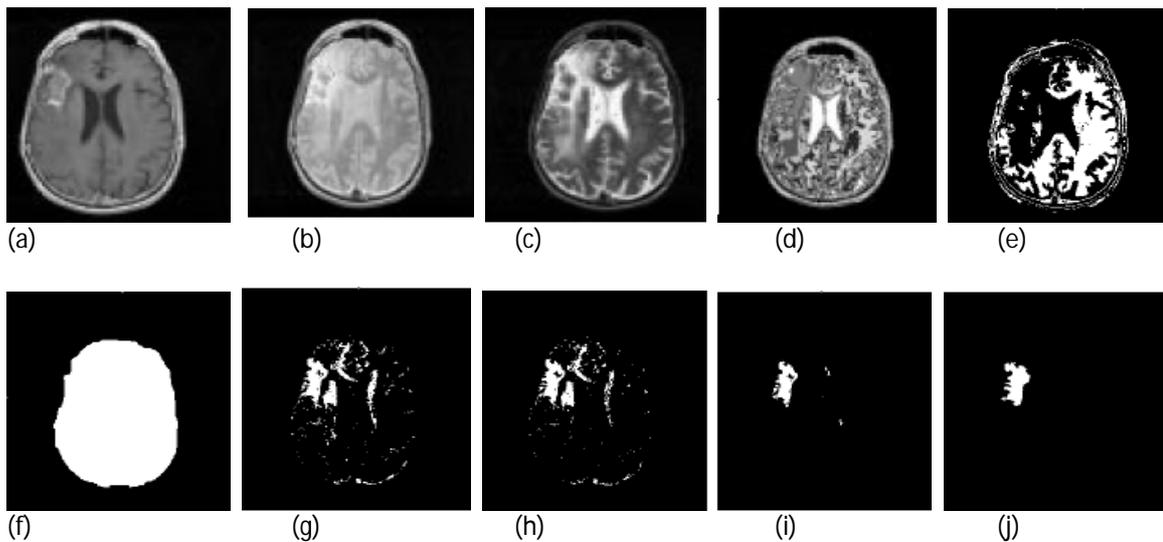

*Figure 19*: Output cited by Matthew C. Clark et. al.[119]. a), b), c)are input MRI T1,PD, and T2 images, d) initial segmentation by unsupervised clustering method, e) white matter class, f) mask created from initial segmentation, g) initial tumor segmentation using adaptive histogram thresholds on mask image, h) tumor segmented using densi screening, i) removal of spatial region that donot contains tumor, j) ground throuth image by expart radiologist.

Disadvantage of this approach is the substantial manual policies requirement, the rules may not be robust to nonstandard intensity and the errors can propagate if the assumptions of early rules in the sequence are not implemented. In the case of heterogeneous tumor or noise in MRI, it may not work correctly.

**7.12 Texture-based:** Texture analysis is a good task in image processing for classification, identification and segmentation of images. Textures are the reproduction, symmetries and amalgamation of large number of basic patterns with some random changes. In texture segmentation the goal is to assign an unknown sample image to one of a set of known texture classes Texture segmentation consist of two phases they are learning phase and recognition phase[120,121]. In the learning phase, target is to build a model or pattern for each the texture content. The texture content of the training images is captured with the selected texture analysis techniques, which yields a set of textural description for each image. These features, which can be scalar numbers or discrete histograms or empirical distributions, or any other type of processing characterize given textural properties of the images, such as spatial structure, contrast, roughness, orientation, brightness, intensity etc. In the recognition phase the texture content of the unknown sample is first described with the same texture analysis method. Then the textural features of the sample are compared to those of the training images with a classification algorithm, and the sample is assigned to the category with the best match. Herlidou-Meme et al., [122](2003) have evaluated the effectiveness of texture analysis to illustrate healthy and pathologic human brain tissues such as white matter, gray matter, cerebrospinal fluid, tumors and edema in a larger data set. Each selected region of interest was categorized by both its mean gray level values and several texture parameters and a multivariate statistical examination was then applied in order to differentiate each brain tissue type signified by its own set of texture parameters. Four statistical texture analysis methods were used these are histogram, co-occurrence matrix, gradient matrix and run-length matrix and they were previously performed on test objects to evaluate the technique confidence on acquisition parameters and consequently the interest of a multicenter evaluation. The results show that there is a relatively good discrimination between the tumor and its surrounding edema but no discrimination was made between solid part and cystic or necrotic parts.

Busch[123] in 1997 presented another texture-based method to segment a definite type of non-enhanced homogeneous tumor in T1-weighted, T2-weighted, and co-registered CT images. This method used five texture extraction methods to compute features. The results of the five classifiers were weighted and combined. Finally a facts based post processing using morphological operations was used to eliminate the misclassified voxels and to refine the result. The use of multiple classifiers certified a extra robust classification than the individual classifiers. Second order textures provided the worst classification performance among the five texture extraction methods.

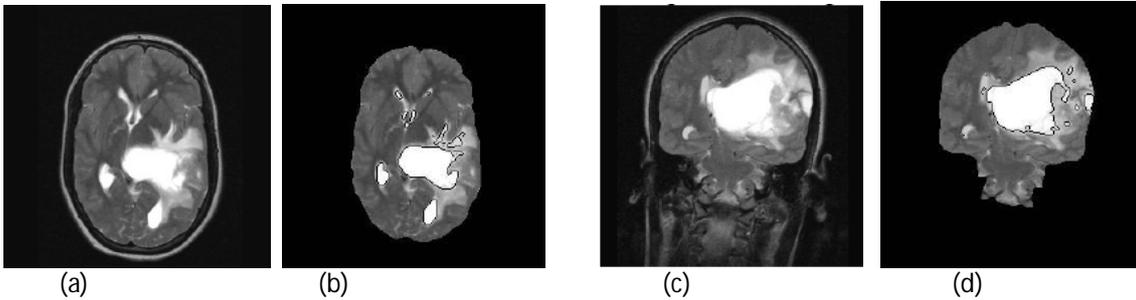

(a)　　　　　　　　　(b)　　　　　　　　　(c)　　　　　　　　　(d)

**Figure 20:** Output cited by Qurat-Ul-Ain et. at. [120] where b) and c) for input a) and b) as MRI of brain image, using texture analysis

Texture based methods need a culture procedure and can segment particular types of tumor. Generalization of these methods to different more types of tumors is too much difficult. In addition it seems that these methods cannot segment all components of the tumor and are sensitive to noise and inhomogeneity.

**7.13 Artificial neural networks (ANNs):** ANN is one of the powerful AI techniques that have the capability to learn from a set of data and construct weight matrices to represent the learning patterns. Artificial neural networks (ANNs) are massively parallel networks of processing elements or nodes that simulate biological learning. Each node in an ANNs is capable of performing elementary computations [125]. The motivation for the development of neural network technology stemmed from the desire to develop an artificial system that could perform intelligent tasks similar to those performed by the human brain.

ANN is a mathematical model which emulates the activity of biological neural networks in the human brain. It consists of two or several layers each one has many interconnected group of neurons[126, 127, 128]. Each neuron in the ANN has a number of inputs vector $P$ and one output $Y$. The input vector elements are multiplied by weights $w_{1,1}, w_{1,2},..., w_{1,R}$, and the weighted values are fed to the summing junction. Their sum is simply the dot product ($W.P$) of the single-row matrix $W$ and the vector $P$. The neuron has a bias $b$, which is summed with the weighted inputs to form the net input $n$. This sum, $n$, is the argument of the transfer function $f$:

$$n = \sum_{i=1}^{R} w_{1,i} \cdot p_i + b,$$

$$Y = f(W \cdot p + b),$$

$$Y(j) = f\left[\sum_{i=1}^{R} w_{1,i}(j) \cdot p_i(j) + b\right]$$

The learning process can be summarized in the following steps: (1) the initial weights are randomly assigned, (2) the neuron is activated by applying inputs vector and desired output ($Yd$), and (3) calculation of the actual output ($Y$) at iteration $j$=1 as illustrated in (3), where iteration $j$ refers to the $j^{th}$ training example presented to the neuron. The following step is to update the weights to obtain the output consistent with the training examples, as illustrated in

$$w_{i,j}(j+1) = w_{1,i}(j) + \Delta w_{1,i}(j)$$

Where $\Delta w_{1,i}(j)$ is the weight correction at iteration $j$. The weight correction is computed by using the delta rule in

$$\Delta w_{1,i}(j) = \alpha \times p_i(j) \times e(j),$$

where $\alpha$ is the learning rate and $e(j)$ is the error which can be given by

$$e(j) = Y_d(j) - Y(j).$$

Finally, the iteration $j$ is increased by one, and the previous two steps are repeated until the convergence is reached. All voxels with intensities $f(x, y)$ larger than the threshold value $T$ are allocated into one class, and all the others into another class.

$$g(x,y) = \begin{cases} f(x,y) & if\ f(x,y) > T \\ 0 & if\ f(x,y) < T \end{cases}$$

Clustering techniques aim to classify each voxel in a volume into the proper cluster, and then these clusters are mapped to display the segmented volume. The most commonly used clustering technique is the *K*means method, which clusters *n* voxels into *K* clusters (*K* less than *n*) Its main objective is to achieve a minimum intracluster variance *V*.

$$V = \sum_{i=1}^{K} \sum_{x_j \in S_j} (x_j - \mu_i)^2$$

Where *K* is the number of clusters, $S = 1, 2, \ldots, K$, and $\mu_i$ is the mean of all voxels in the cluster *i*. K-means approach has been used with other techniques for clustering medical images. Neural networks resemble the human brain in two ways[129]: they acquire knowledge through learning and the knowledge is stored within inter-neuron connection strengths known as synaptic weights. The true power and advantage of neural networks lies in their ability to represent both linear and nonlinear relationships and in their ability to learn these relationships directly from the data being modelled. Traditional linear models are simply inadequate when it comes to modelling data that contains nonlinear characteristics[131]. Because of the many interconnections used in a neural network, spatial information can be easily incorporated into its classification procedures.

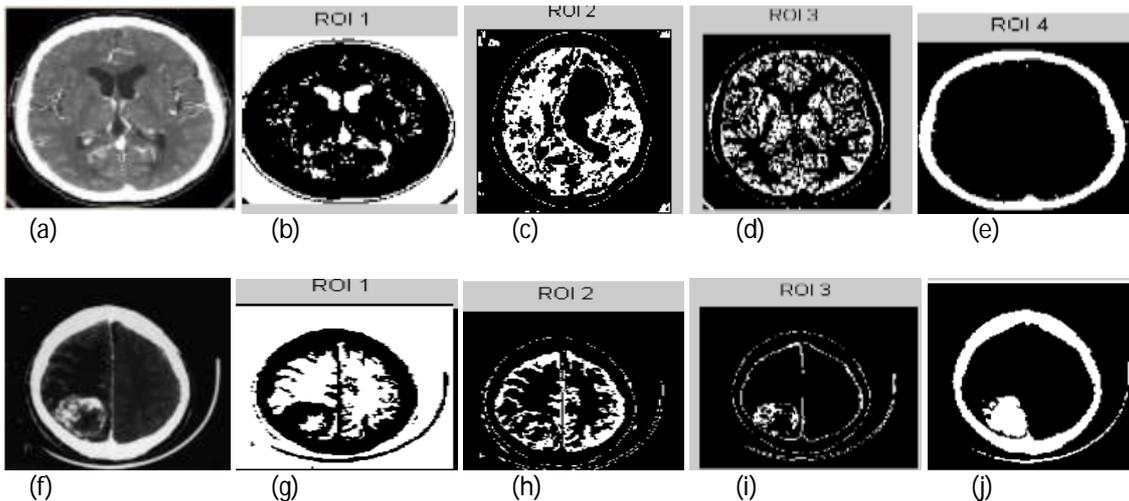

***Figure 21*** : *Output cited by A. Padma [130](2011) for different input using different Artificial neural networks for different region of interest.*

Neural networks execute very well on complicated, difficult, multivariate non-linear domains, such as tumor segmentation where it becomes more difficult to use decision trees, or rule-based systems. They also perform a little better on noisy fields and there is no need to assume a fundamental data allocation such as usually done in statistical modelling. But there are several disadvantages in using neural networks for tumor segmentation. Usually they need a patient-specific learning which a very time consuming process is. Another disadvantage is that neural networks do not give explicit knowledge representation in the form of rules, or some other easily interpretable form. The model is implicit, hidden in the network structure and optimized weights, between the nodes.

**7.14 Fusion-based:** Fusion techniques are based on various theories such as probabilistic and Bayesian fusion, fuzzy set theory, possibility and belief functions theory. Since a tumor consists of different biological tissues, one type of MRI cannot give complete information about abnormal tissues. Therefore, different MRI modalities information of a patient is combined to take a decision on the location, extension, prognosis and diagnosis of the tumors discussed by Ruan et al. (2007)[132]. Data fusion is a growing research field, and the goal of data fusion is to obtain an information synthesis by combining different data [133]. Data coming from different sources and techniques are usually redundant but also complementary.

A fuzzy fusion Dou et al. [134] using operators such as t-norm or average operators was performed to fuse the membership functions. Finally a fuzzy region growing is used to refine the final result. This method uses the fused information of several MRI types to segment the tumor automatically and is very fast to detect and segment the tumors.

**7.15 Fuzzy connectedness:** The Fuzzy Connected Image Segmentation framework developed by [135] assigns fuzzy affinities to the target object during classifications, which are used for the segmentation of the target object in the image. The affinity between the two given pixels in an image is defined as a combined weighted function of the degree of coordinate space adjacency [136], the degree of intensity space adjacency, and the degree of intensity gradient space adjacency to the corresponding target object features and this process is shortly describe by Jayaram K. Udupa et al. mathematically as follows [136, 137, 138, 139, 140]:

A binary scene over a fuzzy digital space $(Z^n, \alpha)$ is a pair $\tau = (C, f)$, where $C$ is a $n$-dimensional array and $f$ is a function whose domain is $C$, called the scene domain, and whose range is a subset between 0 and 1. Fuzzy affinity $k$ is any reflexive and symmetric fuzzy relation in $C$, that is

$$k = \{((c,d), \mu_k(c,d)) \mid (c,d) \in C\}$$
$$\mu_k : C \times C \to [0,1]$$
$$\mu_k(c,c) = 1, \quad \forall c \in C$$
$$\mu_k(c,d) = \mu_k(c,d), \mu_k \forall (c,d) \in C$$

The general form of $\mu_k$ can be written as follows:

$$\mu_k(c,d) = h(\mu_\alpha(c,d), \mu_\beta(c,d), \mu_\gamma(c,d), c, d) \forall (c,d) \in C$$

$\mu_\alpha(c,d)$ Represents the degree of coordinate space adjacency of $c$ and $d$; $\mu_\beta(c,d)$ represents the degree of intensity space adjacency of $c$ and $d$; and $\mu_\gamma(c,d)$ represents the degree of intensity gradient space adjacency of $c$ and $d$ to the corresponding target object features.

Fuzzy $k$-connectedness $K$ is a fuzzy relationship in C, where $\mu_K(c,d)$ is the strength of the strongest path between $c$ and $d$, and the strength of a path is the smallest affinity along the path. A fuzzy connected component is defined as a hard binary relationship $K_\theta$ in $C$ based on the fuzzy $k$- connectedness:

$$\mu_K(c,d) = \begin{cases} 1, & \mu_K(c,d) \geq \theta \in [0,1] \\ 0, & \text{for other} \end{cases}$$

Let $O_\theta$ be an equivalence of the relation $K_\theta$ in $C$. A fuzzy $k$-component $\Gamma_\theta$ of $C$ of strength is a fuzzy subset of $C$ defined by the membership

$$\mu\Gamma_\theta = \begin{cases} f(c), & c \in O_\theta \\ 0, & \text{for other} \end{cases}$$

The equivalence class $O_\theta \in C$, such that for any, $(c,d) \in C$, $\mu_K(c,d) \geq \theta$, $\theta \in [0,1]$ and for $e\{C - O_\theta\}$, $\mu_K(c,d) < \theta$. The notation $[O]_\theta$ is used to denote the equivalence class of $K\mu$ that contains $O$. The fuzzy $k$-component of $C$ that contains $O$, and the membership function is given by

$$\mu\Gamma_{\theta(O)} = \begin{cases} f(c), & c \in [O]_\theta \\ 0, & \text{for other} \end{cases}$$

In a generic implementation of fuzzy connectedness for $c,d \in C : \mu_k(c,d) = h(\mu_\alpha(c,d), f(c), f(d), c, d)$ where $c, d$ are the image locations of the two pixels, $\mu_\alpha(c,d)$ is an adjacency function based on the distance of the two pixels, and $f(c)$ and $f(d)$ are the intensity of pixels *c* and *d*, respectively. In this general form, $\mu_k(c,d)$ is shift-variant. In other words, it is dependent on the location of pixels *c* and *d*. A more specific and shift-invariant definition for a fuzzy affinity was introduced

$$\mu_k(c,d) = \mu_\alpha(c,d)[\omega_1 h_1(f(c),f(d)) + \omega_2 h_2(f(c),f(d))]$$

$$\mu_k(c,d) = 1$$

Where, $\mu_k(c,d)$ is a linear combination of $h_1(f(c), f(d))$ and $h_2(f(c), f(d))$, with $w_1 + w_2 = 1$.

The adjacency function $\mu_\alpha(c,d)$ is assumed to be a hard adjacency relation, such that:

$$\mu_\alpha(c,d) = \begin{cases} 1, & \sqrt{\sum_i (c_i - d_i)^2} \leq 1 \\ 0, & \text{for other} \end{cases}$$

Where $c_i$ $(0 \leq I \leq n)$ are the pixel's coordinates in *n* dimensions. The functions $h_1$ and $h_2$ are Gaussian functions of $\frac{1}{2}(f(c) + f(d))$ and $|f(c) - f(d)|$ respectively, such that:

$$h_1(f(c),f(d)) = e^{-\frac{1}{2}\left[\frac{\frac{1}{2}(f(c)+f(d)) - m_1}{s_1}\right]^2}$$

$$h_2(f(c),f(d)) = e^{-\frac{1}{2}\left[\frac{\frac{1}{2}|f(c)-f(d)| - m_2}{s_2}\right]^2}$$

Where $m_1$ and $s_1$ are the mean intensity and standard deviation of the intensity of the sample region and $m_2$ and $s_2$ are the mean and standard deviation of the gradient of the sample region[141]. When one uses fuzzy connectedness relation directly for image segmentation, the intensity-based information of an object should be embedded in the affinity function and have the great advantage of detection of multi-component object with low execution time [142, 143, 144, 145]. This information involves distribution of intensity and its inhomogeneities which are provided by selection or estimation of a series of parameters.

Although results of segmentation obtained by fuzzy connectedness method with suitable parameter are often good but these methods have several drawbacks, thus some major drawbacks are follows[136, 138, 146, 147,148]: i) The segmentation results strongly depend on the threshold value used for the binarization of μ , ii) The segmentation results strongly depend on the choice of the functions $h_1$ and $h_2$ defining the pixel affinity to the reference pixel, iii) The results are also strongly dependent on the way parameters are defined in the interaction step.

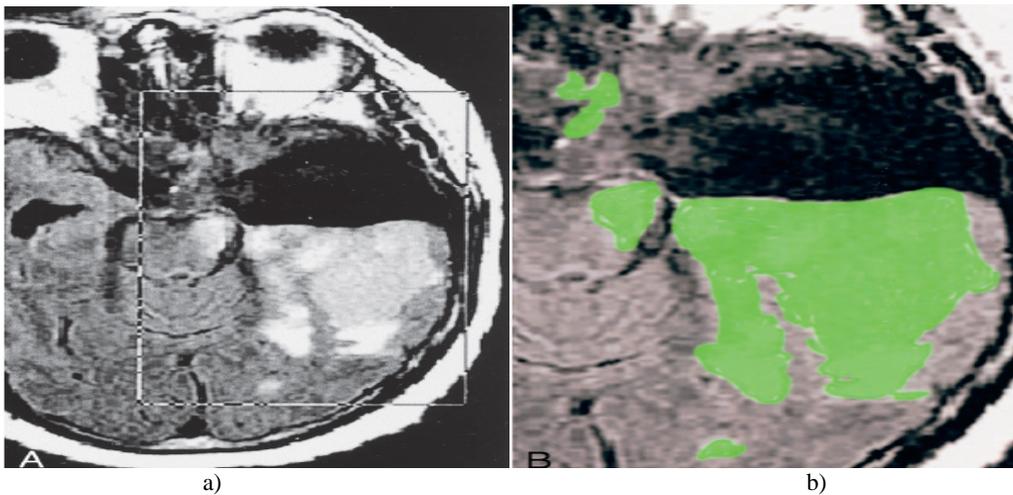

a)      b)

***Figure 22***: *Output cited by Gul Moonis et al. [144] a) Placement of rectangular volume of interest around the area of presumed tumor and edema and b) Delineated fluid-attenuated inversion recovery volume displayed as a green*

To compute a fuzzy-connected object, the strength of connectedness between all possible pairs of voxels in the image must be determined. Fuzzy connectedness has been effectively used to segment out an object in a badly corrupted image. It permits rapid, reliable, consistent, and highly reproducible measurement of tumor volume from MRIs with limited operator interaction. The variability is caused almost entirely by variation in selection of seed points and in editing decisions. Also it needs much calculation time for calculating the connectedness of a path and therefore the algorithm is relatively slow.

**7.16 Watershed Methods:** It is one of the best methods to group pixels of an image on the basis of their intensities. Watershed algorithm is based on morphological process although it can be mixed up with edge based segmentation to yield a hybrid technique. Normally, images acquired by various techniques in the electromagnetic spectrum, possesses a large no of discontinuities in the intensity and these ultimately give rise to over segmentation when morphological segmentations like watersheds are carried out. Pixels falling under similar intensities are grouped together [150]. It is a good segmentation technique for dividing an image to separate a tumor from the image Watershed is a mathematical morphological operating tool.

Mathematically watershed segmentation shortly describes[150,151,153]; Suppose the lower point in the image (water shade and catchment basins ) are $LP_1$, $LP_2$.....$LP_Z$ to be coordinate of these points for the image I (i, j) and $UB_m$ refers to the points of catchment basins associated with minimum region $LB_z$ (x, y) represented by X[n] accordingly I(x, y) < n.

$$X[n] = \{(x, y)\} \mid I(x, y) < n\}$$

X[n] is the coordinate of points in I (i, j) geometrically lying under the plain I (i, j) =n. Topographically the image filled with water in integer filling increments begin from $n=t_{max-1}$ to $n = t_{min+1}$. Min and max, is the minimum and maximum gray level value. The number of points under the fluid is necessary, due to that marker will used in black color for the coordinates in X[n] which are below the level I(i,j)=n and the other point in white color.

$$C_n B_m = UP_m \cap X[n]$$

Where $C_n B_m$ represents the coordinates in catchment basins related to $P_m$, which are fluid filling at the level n. Then let C[n] refer to the union of the filling fluid of the points of catchment basins of level n:

$$C[n] = \bigcup_{m=1}^{z} B_m$$

Finally $C_{tmax+1}$ refer to all catchment basins union.

$$C_{tmax+1} = \bigcup_{m=1}^{z} UP_m$$

Thus from above equations C[n] is a subgroup of X[n] and watershed lines is prepared when C[tmin+1] =X [tmin+1]. Then the procedure follows to reconstruct C[n-1] at level n. C[n] can be obtained from C[n-1] by assuming S as the set of the connected component in X[n], at s ∈ S[n] there are three assumption i) s ∩ C[n-1] is empty , ii) s ∩ C [n-1] contains more than one connected component of C[n-1] and this lead to s is incorporated in to C[n-1] to produce C[n], iii) In last case s ∩ C[n-1] contains more than one connected component of C[n-1].

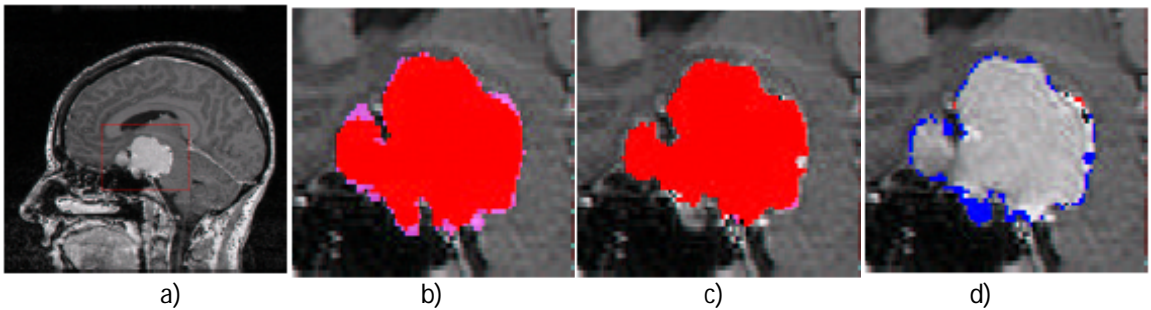

a)     b)     c)     d)

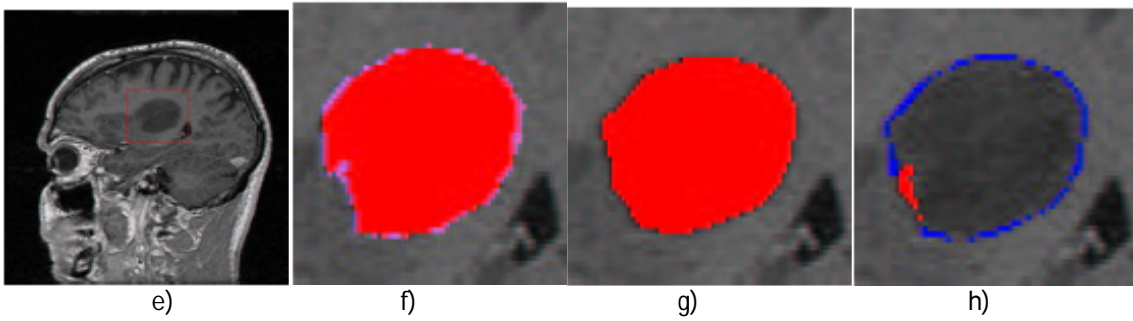

*Figure 23: Output cited by Joshua E. Cates et al.[150] (2003): a) and e) are original image, b) and f) are the manual segmentation by radiologist, c) and g) are the watershed output, d) and h) are the difference of radiologist output and watershed output.*

Watershed Techniques have the ability to detect the continuous boundary of the region of interest, it can be best suited for those types of applications where high accuracy and precision is needed. Detection of tumor in the area of cancer research is best suited area of application where watershed segmentation can be applied efficiently. Watershed is a gradient-based segmentation technique where different gradient values are considered as different heights. A hole is made in each local minimum and immersed in water; the water will rise until local maximums. When two body of water meet, a dam is built between them. The water rises gradually until all points in the map are immersed. The image gets segmented by the dams. The dams are called watersheds and the segmented regions are called catchments basins [154]. The main problem of watershed transform is its sensitivity to intensity variations, resulting in over segmentation, which occurs when the image is segmented into an unnecessarily large number of regions. The over segmentation problem still exists in this method [154].

**7.17 Level Set Methods:** Level set methods use non parametric deformable models with active contour energy minimization techniques which solve computation of geodesics or minimal distance curves. Level set methods are governed by curvature defining speeds of moving curves or fronts.

There are large numbers of level set methods developed for segmentation of medical images and all most all these methods follow some common generic steps [155]. First placement of an initial contour arbitrarily, outside or inside the region of interest, level set ϕ = signed Euclidean distance function of the contour and Function ϕ allowed to evolving according to first or second derivative partial differential equation (PDE), then it is reinitialized after a number of iterations and go to second statement until the function $\varphi$ converges or = 0.

The placement of the initial contour is still a key challenge in some level set segmentation methods. The contour can move inward or outward and its initial placement determines the segmentation target [156]. In some methods, such as outlined in [157, 158], the initial contour is replaced with a region-based contour and the re-initialization step has been eliminated by including a term in the PDE that penalizes the deviation of level set function from a signed distance function. Different level set methods differ either in terms of their initial contour or the energy functional to be minimized or some combination of both. There are still key challenges in this area and there is no general level set method that works for all applications. For different applications, a number of PDEs can be used and the solutions of PDEs are susceptible to the choice of the parameters appearing in the energy functional.

Level sets methods rely on two central embeddings; first the inserting the interface as the zero level set of a higher dimensional function, and secondly the interface's velocity to this higher dimensional level set function. The development of the contour or surface is managed by a level set equation. The solution tended to by this partial differential equation is computed iteratively by updating ϕ at each time interval, the general form of the level set equation is shown below.

$$\frac{\partial \phi}{\partial t} = -|\nabla| F$$

The evolving contour deforms with a speed F that is based on the contour curvature and image features like gradient. The curvature component in the speed keeps the contour propagating smoothly, which performs like the internal energy in snakes. Additionally, an artificial speed term, obtained from the image feature, is synthesized to stop the front, i.e., the contour, at the desired boundary.

In the above level set equation *F* is the velocity term that describes the level set evolution. By manipulating *F*, we can guide the level set to different areas or shapes, given a particular initialization of the level set function. In [159] *F* is dependent on data and curvature functions only for the purposes of image segmentation. Therefore, we adopt the same methodology making the level set equation take the form

$$\frac{\partial \phi}{\partial t} = -|\nabla| \left[ \alpha D(I) + (1 - \alpha) \nabla \cdot \frac{\nabla \phi}{|\nabla \phi|} \right]$$

Where the data function $D(I)$ tends the solution towards targeted features, and the mean curvature term $\frac{\nabla \cdot \nabla \phi}{|\nabla \phi|}$ keeps the level set function smooth. Weighting between these two is $\alpha \in [0,1]$ a free parameter that is set beforehand to control how smooth the contour should be. The data function $D(I)$ acts as the principal force that drives the segmentation. By making $D$ positive in desired regions or negative in undesired regions, the model will tend towards the segmentation sought after. A simple speed function that fulfills this purpose, used by Lefohn, Whitaker and Cates in [159,160], is given by

$$D(I) = \epsilon - |I - T|$$

Here $T$ describes the central intensity value of the region to be segmented, and $\epsilon$ describes the intensity deviation around T that is part of the desired segmentation. Three user parameters that need to be specified for segmentation are $T$, $\epsilon$ and $\alpha$. An initial mask for the level set function is also required, which may take the form of a square in two dimensions, or any other arbitrary closed shape. The level set iteration can be terminated once α has converged, or after a certain number of iterations.

The motion of the front is matched with the zero level set of a signed distance function, and the resulting partial differential equation for the evolution of the level set function resembles a Hamilton–Jacobi equation. This equation is solved using entropy-satisfying schemes borrowed from the numerical solution of hyperbolic conservation laws, which enable the topological changes, corner and cusp development to be naturally obtained during the front marching process [161].

$$\frac{\partial \phi}{\partial t} = F|\nabla|, \phi(x,y,t=0) = \pm d$$

F is the contour marching velocity $F = g(|G_{\sigma_0} * I|)(c+k)$ where $g(|G_{\sigma_0} * I|) = \frac{1}{(1+|G_{\sigma_0} * I|^2)}$, $c > 0$ is a constant, k is the contour curvature, d is the distance from (x,y) to the initial contour, and φ is positive or negative when (x, y) is either outside or inside the initial contour. To speed up the front marching, computationally efficient schemes like the narrow-band method and the fast marching method are proposed.

Based on the above level set methods, a new level set method called geodesic active contour method was proposed which unified the curve evolution approaches with the classical energy minimization methods (snakes). The flow of the contour can be described as

$$\frac{\partial \phi}{\partial t} = g(|G_{\sigma_0} * I|)(c+k)N' - (\nabla g \cdot N')N'$$

And $g(|G_{\sigma_0} * I|) = \frac{1}{(1+|G_{\sigma_0} * I|^2)}$, $c > 0$ is a constant for fast convergence and k is the curvature. The embedding surface deformation process using level set implementation is

$$\frac{\partial \varphi}{\partial t} = g(|G_{\sigma_0} * I|)(c+k)|\nabla \varphi| + \nabla \varphi \cdot \nabla \varphi$$

With g(.), k and $c$ being defined as above. N. S. Zulpe et. al.[155] used basic level set to segment the tumor from the Metastatic bronchogenic carcinoma; MRI scans of 42-year-old woman from the Whole Brain Atlas. They try to find out the contour around the tumor area in 2D MRI images. To specify initial contour we specified the arbitrary ten seed points around tumor, after final iteration,

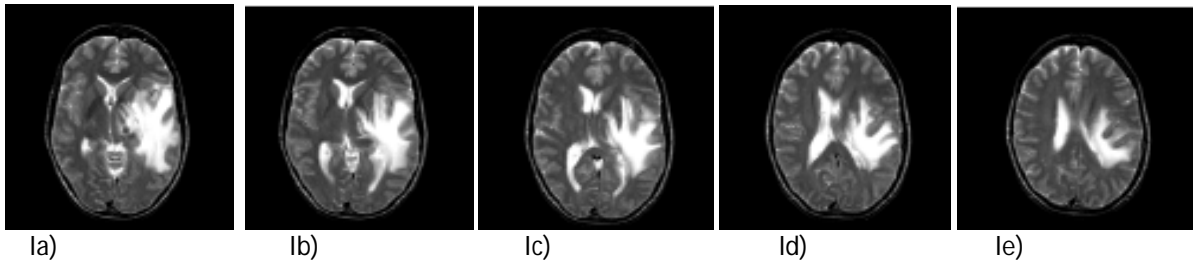

Ia)       Ib)       Ic)       Id)       Ie)

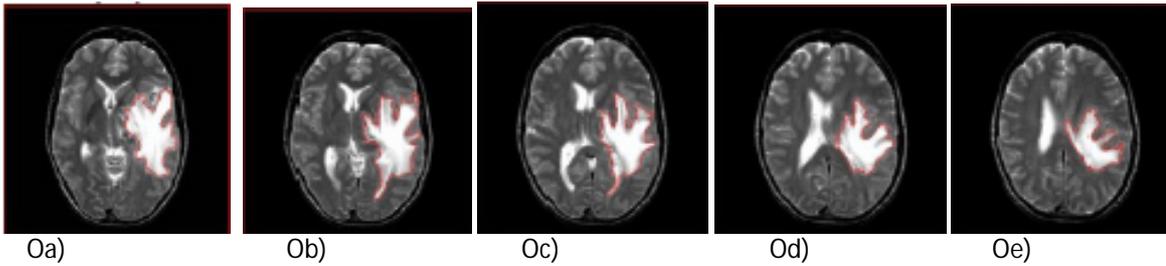

| Oa) | Ob) | Oc) | Od) | Oe) |

*Figure 24: Output cited by the tumor segmentation by N.S.Zulpe (2012)[155] et. al. (Ia-Ie) original image, (Oa-Oe) level-set segmented image.*

Level sets have established a huge prospective for 3D medical image segmentation but their utility has been restricted by two problems[161]. First, three dimensional level sets are comparatively slow to compute. Second, their formulation usually entails several extra or free parameters, which can be very complicated to accurately tune for explicit applications [162]. Thus, level set segmentation is not suitable for the segmentation of complex medical images and they must be combined with powerful initialization techniques to produce successful segmentation.

A possible method to fix this problem is to use the background estimation method based on motion detection techniques using only two images; that is, two images at different times with the pot, pot holder and conveyor mechanism appearing relatively on the same position on the images [163, 164]. Once the background is learned by this method, the background image (with all non objects image) is subtracted from the original image and non-objects image.

Level set methods present a commanding approach for the medical image segmentation because it can handle any of the cavities, concavities, convolution, splitting, or merging. However, this method needs identifying initial curves and can only provide superior results if these curves are placed near symmetrically with respect to the object boundary.

**7.19 The Combination of Watershed and Level Set :** This approach combines the advantages of both methods: the watershed transform pre-segmentation is rough but quick and the level set needs only a few iterations to produce the final, fast, highly accurate, and smooth segmentation. The choice of watershed segmentation as the initialization of the level set method is made for two reasons [166]. The first reason is that because of watershed transform blindness of segmentation is reduced and the accuracy of segmentation is improved. The second reason is to do with improving the computation speed. After the initial segmentation based on watershed transform, the final segmentation is accomplished based on level set method [167]. By combining watershed transform and level sets, this method is able to produce highly accurate segmentations of topologically and geometrically complex structures in much less time than where level sets alone.

**7.18 Self-organizing maps (SOM) :** SOM consists of two layers: first is the input layer and the number of neurons in this layer is equal to dimension of input and second is the competitive layer and each neuron in this layer corresponds to one class or pattern[168]. The number of neurons in this layer depends on the number of clusters and is arranged in regular geometric mesh structure. Each connection from input layer to a neuron in competitive layer is assigned with a weight vector. The SOM functions in two steps, viz, [169] firstly finding the winning neuron i.e. the most similar neuron to input by a similarity factor like Euclidean distance, and secondly, updating the weight of winning neuron and its neighbour pixels based on input.

The basic SOFM model consists of two layers. The first layer contains the input nodes and the second one contains the output nodes. The output nodes are arranged in a two dimensional grid as shown in Figure below

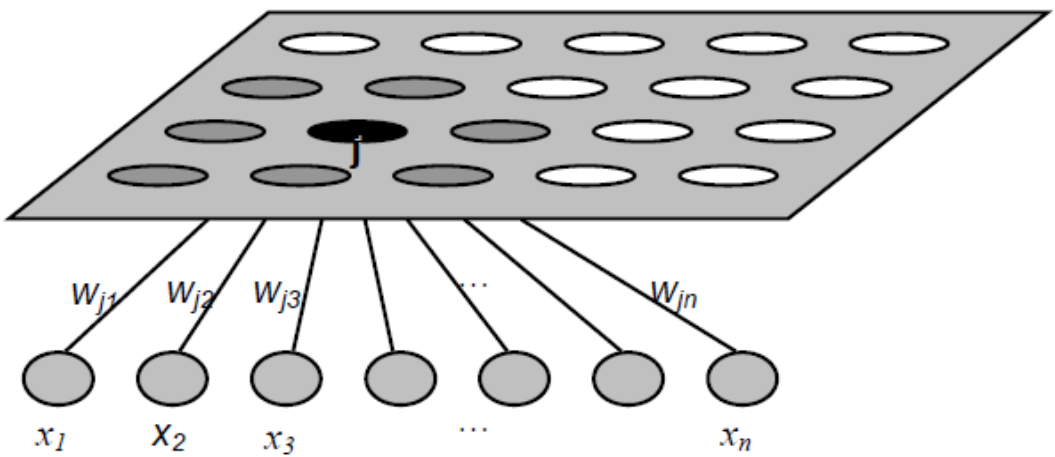

*Figure 25: The basic structure of Self-Organizing Map Network copied from Yan Li et al [170].*

Where upper plane are 2D array of neurons, $w_{ij}$ are Weight, $x_1, x_2,....,x_n$ are Set of input signals, Every input is connected extensively to every output node via adjustable weights. The neighborhood is centered on the output node whose distance $d_{ij}$ is minimum. The measurement of $d_{ij}$ is an Euclidean distance, defined as:

$$d_{ij} = min \|x_i - w_{ij}\|^2$$

The neighborhood decreases in size with time until only a single node is inside its bounds. A learning rate, $\alpha_{ij}(t)$, is also required which decreases monotonically in time. The weight updating rule is as follows:

$$w_{ij}(t+1) - w_{ij}(t) + \alpha_{ij}(t)\left(x_i - w_{ij}(t)\right)$$

SOFM algorithms are, firstly, highly dependent on the training data representatives and the initialization of the connection weights [171,172]. Secondly, they are very computationally expensive since as the dimensions of the data increases, dimension reduction visualization techniques become more important, but unfortunately the time to compute them also increases.

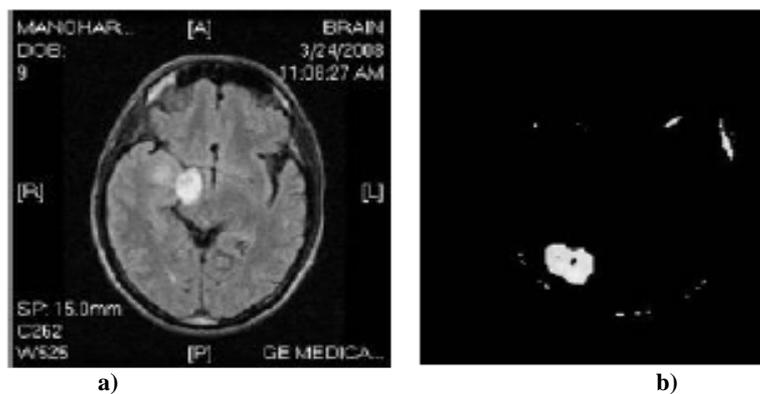

a)     b)

**Figure 26**: Output cited by Vanitha.L et al. [173], a) input MRI of brain image, b) output image by applying SOM methods 1012.

Self-organizing maps (SOM) is an unsupervised clustering network that maps inputs which can be high dimensional to one or two dimensional discrete lattice of neuron units [169]. The input data is organized into several patterns according to a similarity factor like Euclidean distance and each pattern assigns to a neuron. Each neuron has a weight that depends on the pattern assigned to that neuron [169]. Input data is classified according to their grouping in input space and neighbouring neuron and moreover learns distribution and topology of input data [174,175]. For calculating that black and white similarity map, the more neighbors it use to calculate the distance the better similarity map we will get, but the number of distances the algorithm needs to compute increases exponentially.

**7.19 Hybrid SOM:** HSOM combines self organization and topographic mapping technique. HSOM combines the idea of regarding the image segmentation process as one of data abstraction where the segmented image is the final domain independent abstraction of the input image [176, 177, 178]. The HSOM is organized in a pyramidal mannered structure consisting of multiple layers where each layer resembles the single layer SOM. Learning process has sequential corrections of the vectors representing neurons. On every step of the learning process a random vector is chosen from the initial data set and then the best-matching neuron coefficient vector is identified [179]. The most similar to the input vector is selected as a winner.

**7.20 Graph cut based:** Numerous graph techniques are existed which are exploited in image segmentation such as minimum spanning trees, shortest path, graph-cuts etc. Among all these typical graph partitioning methods graph-cuts are comparatively new and the most powerful one for image segmentation by M. Sonka, et al. [180] gives a cut in the graph isolates the source from the sink points connected to the sink are labeled as tumor and points connected to the source as brain. Image segmentation relates basically background and object which can be employed as binary labeling problem. Boykov and Jolly [181] in 2001 mentioned the segmentation of a monochrome image that solves a two labels problem in the graph cut method. Considering a set of labels $L$ and a set of sites $S$, the labeling problem can be assigned as a label $f_p \epsilon L$ and each of the site $p \epsilon S$. The label set $L = \{0,1\}$ where 0 indicates background and 1 indicates object. For a labeling problem if $f = f_p | f_p \epsilon L$ for all pixels, the energy minimization Markov Random Field (MRF) equation [182] can be written as:

$$E(f) = \sum_{p \epsilon S} D_p(f_p) + \lambda \sum_{(p,q) \epsilon N} \omega_{pq}. T(f_p \neq f_q)$$

In the energy minimization equation, the first term called as data term consists of constraints from the observed data and measures how the labels are assigned. Label $f_p$ fits with site p and is measured by $D_p$. The second term which is the smoothness term measures to what extent f is not piecewise smooth. $N$ represents the neighborhood system like 4 or 8-connected system. If $f_p = f_q, T(f_p \neq f_q)$ becomes 0 and 1 otherwise. In image segmentation it is expected the boundary to be positioned on the edges. Hence the typical selection of $\omega_{pq}$ is:

$$\omega_{pq} = \varepsilon^{-\frac{(I_p - I_q)^2}{2\delta^2}} \cdot \frac{1}{dist(p,q)}$$

Color values of Sites p and q are represented by $I_p$ and $I_q$ along with distance between p and q is presented by *dist (p,q)*. Level of variation between neighboring sites is expressed by the parameter $\delta$. The relative importance of the data term versus smoothness term is revealed by the parameter $\lambda$. The graph cuts method is aimed to minimize the objective function i.e. energy of the image corresponding all required labeling for the object and background seeds. The summarization of steps used in graph cuts method can be depicted as below. i. An edge directed graph representing size and dimension of the target segmenting image has to be created [183, 184]. ii. Object and background seeds have to be distinguished properly with formation of two graph nodes-source s and sink t. Based on the object or the background labels, all seeds have to be connected with either source or sinks node. iii. According to table 1 each link of the formed graph is to be associated with suitable edge cost. iv. Any minimum s-t cut method is to be used which indicates the graph nodes representing image boundaries for object and background. v. A suitable maximum flow solution for graph optimization is to be determined for graph cuts segmentation. Graph-based segmentation is one of the commonly used techniques in imaging analysis. Often image segmentation is compared to graph partition procedures with the terminologies- source node i, sink node j, capacity based weight matrix $W_{ij}$.

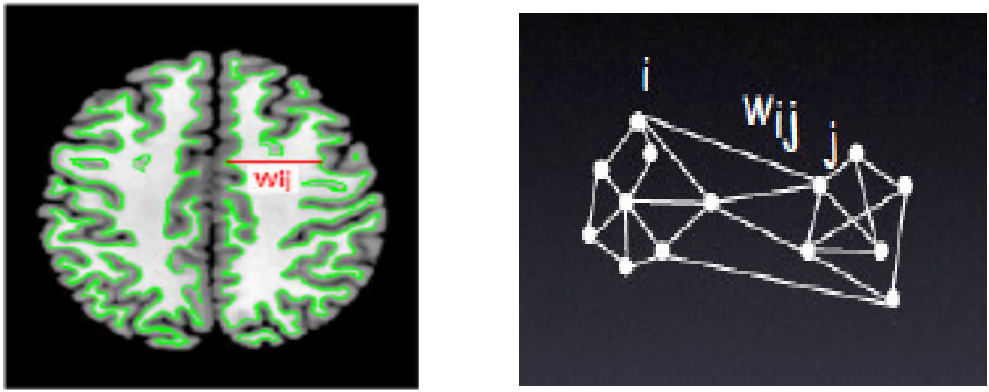

*Figure 27: Graph formation from an image, I = {pixels} having source, sink and edge connection nodes: the figure is cited from the tutorial used in Mohammad Shajib Khadem[184].*

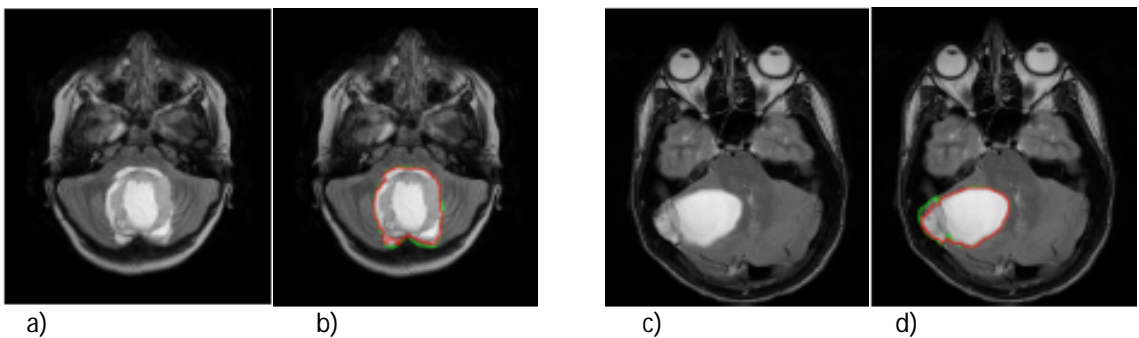

a)    b)    c)    d)

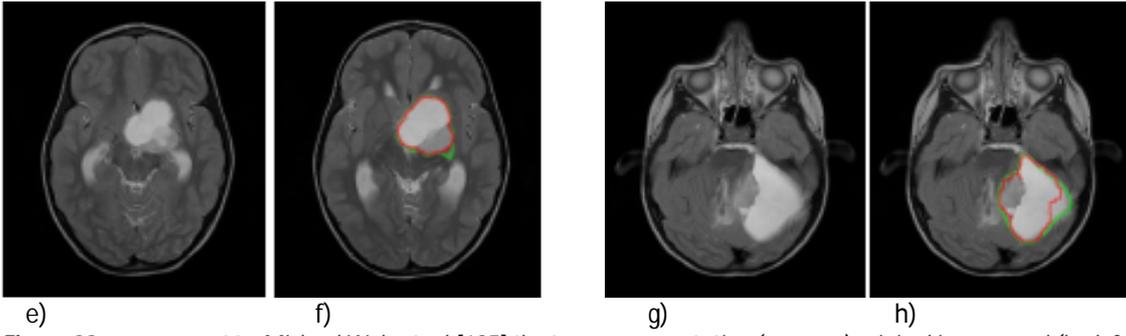

e)  f)  g)  h)

**Figure 28**: *Output cited by Michael Wels et. al.[185] the tumor segmentation (a, c, e, g) original images and (b, d, f, h) corresponding graph cut based segmented image.*

Here, the problem of image segmentation is considered as a graph partitioning problem and global criterion that measures both total dissimilarity among the different groups and the total similarity inside then is used. An efficient method based on generalized Eigen value treatment is used to optimize the criterion to segment image.

**7.21 Fractal-based:** A fractal is an irregular geometric object with an infinite nesting of structures at all scales. Some of the most important properties of fractals are self-similarity, chaos, and non-integer fractal dimension (FD)[186, 187, 188]. Mathematically, a fractal structure is defined as a set that has a fractal dimension exceeding its topological one. FD serves as an index of the morph metric complexity and variability of the object being studied. The disadvantage is that the size of sub images is a problem, because different sub image sizes result in different FD. The second problem is the selection of reference images, because the MR images have different sizes and different parameters [189] and for tumor detection it is required to have a reference image similar to the patient image.

**7.22 Parametric deformable models (snakes):** Parametric models explicitly move predefined snake points based on an energy minimization scheme [190]. The following section reviews available segmentation algorithms on parametric methods. The deformation process has played a critical role in shape representation. The first class of deformable model is parametric deformable curves model, also known as snakes. Since then, there has been an extensive burst of publications in the area of parametric deformable models and their improvements, such as balloon force, topology snake, and distance snake.

In 2-D, a snake is defined as a curve C(s) =x(s), y(s) where x∈ (0, 1) in traditional snake, the energy usually formed by internal forces and external forces[191] as,

$$E_{snake} = E_{internal} + E_{external}$$

Where $E_{internal}$ and $E_{external}$ are internal and external energies, respectively. The internal energy function determines the regularity, i.e., smooth shape of the contour. This energy is given by

$$E_{internal} = \int_0^1 \alpha |c'(s)|^2 + \beta |c''(s)|^2 ds$$

Here, α controls the tension of the contour and β control the rigidity of the contour. C'(s) and C''(s) represents first and second derivatives, respectively[192]. External energy, $E_{external}$ intends to pull or push the curve towards the edges. Typically, the external energy consists of potential forces. This energy is generally the image force and is defined as

$$E_{external} = \int_0^1 E_{image}(C(s))ds$$

Where $E_{image}$ represents the negative gradient of potential function defined on the image plane, so that local minimum of $E_{image}$ attracts the snakes to edges. This edge attraction function is a function of image gradient and is defined as,

$$E_{image}(x,y) = \frac{1}{\lambda |\nabla G_\sigma * I(x,y)|}$$

where G denotes a Gaussian smoothing filter with standard deviation σ, $\lambda$ is a suitably chosen constant, and '*' is a convolution operator. When the snake moves through the spatial domain of an image to minimize the energy functional,

$$E_{snake} = \int_0^1 \alpha |c'(s)|^2 + \beta |c''(s)|^2 + E_{external} C(s) ds$$

The internal energy function determines the regularity, i.e., smooth shape of the contour. Here $\alpha$ controls the tension of the contour, and $\beta$ controls the rigidity of the contour. Application of this basic model is limited because of contour initialization. The internal energy function determines the regularity, i.e., smooth shape of the contour. Here $\alpha$ controls the tension of the contour, and $\beta$ controls the rigidity of the contour. Application of this basic model is limited because of contour initialization. Many variations, extension, and alternative formulations appeared since the introduction of traditional snake model [193]. Berger [194] has proposed the first and primary uses of parametric models in medical image analysis to segment objects in 2D images. However, this classic snake model provides an accurate location of the edges only if the initial contour is given sufficiently near the edges, because they make use of only the local information along the contour.

This limitation indicates that, basic snake model alone cannot serve the purpose of accurate segmentation and they need further modifications and extensions. original snake model [193] and the contour curve is treated as a balloon that is inflated in order to avoid local minima solutions i.e., the curve passes over edges and is stopped only if the edge is strong. However, it does not work image with weak edges. Xu and Prince [194] have made an effort by introducing gradient vector flow as an external force (region based features), which significantly increases the capture range. In this method, they replaced the potential force in the traditional equation with a novel external force field called Gradient Vector Flow (GVF). In order to build up this field, an appropriate edge map function f(x,y) having larger value near the edge is chosen. It is a 2-D vector field that minimizes the following objective function,

$$E(C) = \iint \alpha(u_x^2 + u_y^2 + v_x^2 + v_y^2) + |\nabla f|^2 |V - \nabla f|^2 dx dy$$

Where V=(u(x,y), v(x,y)) and $u_x$, $u_y$, $v_x$, $v_y$, are the spatial derivatives of the field, and $\nabla f$ is the gradient of the edge map, which is defined as the negative external force, i.e., $f=-E_{external}$. This method makes [195] the model free from the initial conditions and also they can handle concave objects. But still it poses the following drawbacks. i. Since the boundary information is not used directly, strong edges as well as weak edges create a similar flow due to the diffusion of the flow information. ii. The generation of GVF is iterative and computationally intensive. Paragios et. al. [196] have introduced a set of diffusion equations that is applied to image gradient vectors yielding a vector field over the image domain.

$$\frac{dC}{dt} = g(k + Vk(x)) + (1 - |k(x)|)[u,v]N'$$

Where u and v are the normalized GVF; this method has the bidirectional flow. This method can extract concave object extraction problem, however, it suffers from high computational requirements. To overcome these limitations, Cvancarova et al. [197] have introduced several improvements to the original GVF algorithm. These modifications are as follows. i. Coupling the smoothness of the edge map to the initial size of the snake ii. Tuning the regulating parameters of the snake iii. Improving the numerical approximation of a vector diffusion equation of the GVF Traditional snake often converges to local minimum of equation and they do not perform well on noisy images, and their capture range is small. This problem is overcome by Osher and Sethian [198]; they suggested some external force model to enhance the capture range. The edge map is computed by first smoothing with a Gaussian kernel followed by a gradient operator or Gabor filters to enhance the tumor boundaries. In addition to computing the edge map, they have also modified the GVF algorithm. By introducing several parameters to the GVF algorithm, the capture range is shown to be satisfactory. However, this method requires the prior knowledge of the object and the parameter selection depending on the initialization to achieve good accuracy.

Mathematically, a deformable contour is a curve X which moves through the spatial domain of an image to minimize this energy function: $E(X) = F_{int}(X) + F_{ext}(X)$ where $F_{int}$ is the internal force that constrains the regularity of the curve and $F_{ext}$ is the external force. The internal force is usually defined as: $F_{int} = \alpha \nabla^2 X - \beta \nabla^2(\nabla^2 X)$ where α and β respectively control the curve tension and rigidity. Deformable models are model-based techniques for delineating region boundaries by using closed parametric curves or surfaces that deform under the influence of internal and external forces. To delineate an object boundary in an image, a closed curve or surface must first be placed near the desired boundary and then allowed to undergo an iterative relaxation process. Internal forces are computed from within the curve or surface to keep it smooth throughout the deformation. Deformable models are extensively used in the segmentation of medical images.

Parametric contour based methods detect tumor boundaries better than region based methods but they have two main limitations. First, when the initial model and the desired object boundary differ largely in size and shape, the model must be reparameterized to fully recover the object boundary. The second limitation with the parametric approach is that it has difficulty dealing with topological adaptation such as splitting or merging model parts.

**7.24 Edge-based segmentation methods:** In this method an algorithm searches for pixels with high gradient values that are usually edge pixels and then tries to connect them to produce a curve which represents a boundary of the object. The user determines an initial guess for the contour, which is then deformed by image driven forces to the boundaries of the desired objects. In these models, two types of forces are considered. The internal forces, defined within the curve, are designed to keep the model smooth during the deformation process. The external forces, which are computed from the image data, are defined to move the model toward an object boundary. The Canny edge detection algorithm [199] is known to many as the optimal edge detector. Canny's intentions were to enhance the many edge detectors already out at the time he started his work. Canny has shown that the first derivative of the Gaussian closely approximates the operator that optimizes the product of signal-to-noise ratio and localization. The algorithm then tracks along these regions and suppresses any pixel that is not at the maximum. The steps for canny edge detection is follows, Compute $f_x$ and $f_y$

$$f_x = \frac{\partial}{\partial x}(f * G) = f * \frac{\partial}{\partial x}G = f * G_x$$
$$f_y = \frac{\partial}{\partial y}(f * G) = f * \frac{\partial}{\partial y}G = f * G_y$$

$G(x, y)$ is the Gaussian function; $G_x(x, y)$ is the derivate of $G(x, y)$ with respect to x:

$$G_x(x, y) = \frac{-x}{\sigma^2} G(x, y)$$

$G_y(x, y)$ is the derivate of $G(x, y)$ with respect to y:
$$G_y(x, y) = \frac{-y}{\sigma^2} G(x, y)$$

The performance of the canny algorithm [199] depends heavily on the adjustable parameters, σ, which is the standard deviation for the Gaussian filter, and the threshold values, 'T1' and 'T2'. σ also controls the size of the Gaussian filter. The bigger the value for σ, the larger the size of the Gaussian filter becomes. This implies more blurring, necessary for noisy images, as well as detecting larger edges. Gradient-based algorithms such as the Prewitt filter have a major drawback of being very sensitive to noise. The gradient is a vector which has certain magnitude and direction is define as follows, the magnitude of gradient provides information about the strength of the edge and the direction of gradient is always perpendicular to the direction of the edge is the concept of sobel operator[25].

$$\nabla f = \begin{pmatrix} \partial f / \partial x \\ \partial F / \partial y \end{pmatrix}$$

These kernels are designed to respond maximally to edges running vertically and horizontally relative to the pixel grid, one kernel for each of the two perpendicular orientations. The kernels can be applied separately to the input image, to produce separate measurements of the gradient component in each orientation. These can then be combined together to find the absolute magnitude of the gradient at each point and the orientation of that gradient. The gradient magnitude is given by:

$$mag(\nabla f) = \sqrt{(\partial f / \partial x)^2 + (\partial f / \partial x)^2} = \sqrt{M_x^2 + M_y^2}$$

The angle of orientation of the edge giving rise to the spatial gradient is given by:

$$dir(\nabla f) = \tan^{-1}\left(M_x / M_y\right)$$

Faster to compute magnitude can be also being written as:
$$mag(\nabla f) = |M_x| + |M_y|$$

*Prewitt edge-finding filter* Prewitt operator[200] is similar to the Sobel operator and is used for detecting vertical and horizontal edges in images and enhances the tumor tissue significantly and removes the image edges vigorously and moves the vertices in the direction of the boundaries of the desired structure.
 The Roberts Cross operator [25, 201]   performs a simple, quick to compute, 2-D spatial gradient measurement on an image. These can then be combined together to find the absolute magnitude of the gradient at each point and the orientation of that gradient.

The angle of orientation of the edge giving rise to the spatial gradient (relative to the pixel grid orientation) is given by:

$$\operatorname{dir}(\nabla f) = \tan^{-1}\left(M_x/M_y\right) - 3\pi/4$$

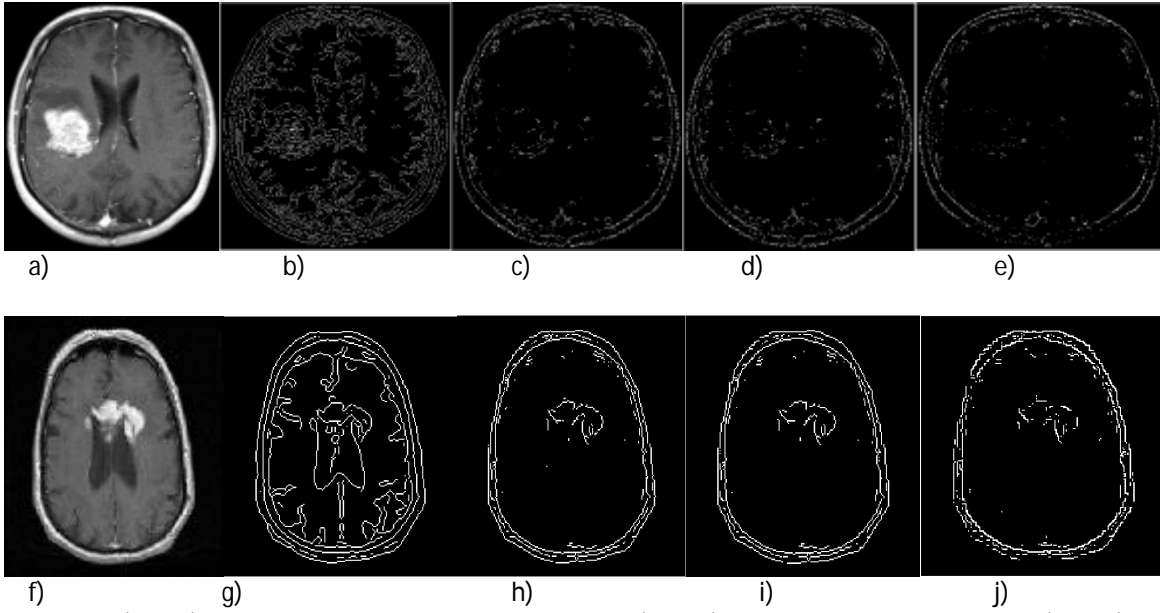

a)   b)   c)   d)   e)

f)   g)   h)   i)   j)

*Figure 29*: a) and f) are original MRI image with brain tumor, b) and g) are corresponding Canny edge, c) and h) are corresponding Sobel edge, d) and i) are corresponding Prewitts edge, e)and j) corresponding Robert edge.

**7. 25 Geometric deformable model:** Deformable models are physically motivated, model based techniques for delineating region boundaries using closed parametric curves or surfaces that deform under the influence of internal and external forces. Deformable models, including active contours (2-D) and active surfaces (3-D) are artificial, closed contours/surfaces able to expand or contract over time, within an image and confirm to specific image features. The following section reviews various types of deformable models used for the segmentation of medical images.

Geometric models move contours implicitly as a particular level of a function. Although several modifications [206] and extensions have been proposed in parametric models, some of the issues such as topological changes and stable convergence remain unsolved. To overcome these limitations geometric models are proposed. Caselles et al. [207, 208] and Malladi et al. [209] have introduced a new method called Geometric Active Contour model. These models can handle topology changes without any additional task.

This method overcomes the limitations of parametric models in capturing multiple objects. However, if there are weak edges or gaps along edges, the evolving curve tends to pass through these areas because gradient values are not large enough for g(x) to approach zero. Caselles et al. [208] have proposed another geometric snake using energy to search for a curve of minimal weighted length called Geodesic Active Contour model, which is proven to be a special case of the traditional snake model. Consider the traditional snake energy equation and substituting $\beta = 0$ then the energy minimization functional for geodesic active contour is given by

$$E(C)_{geo} = \int_0^1 \left(\frac{1}{2}\alpha|C'(s)|^2 - \lambda|\nabla I(C(s))|\right) ds$$

This can be generalized by replacing the edge detector – $|\nabla I|$ with a decreasing function $g|\nabla I|^2$ then the evolution equation for geodesic active contour is

$$\frac{d\varphi}{dt_{geo}} = g.(k + v).|\nabla\varphi| + \nabla g.\nabla\varphi$$

Comparing this with the Geometric Active Contour equation, it is observed that the extra stopping term is used to increase the attraction of the evolving contour towards an edge, of special help when the edge has different gradient values or gaps along it.

Sapiro et al. [210] have also proposed the modified geometric active contour model by introducing the curvature term in the original snake model as,

$$\frac{dE(C)}{dt} = \int_0^1 \left[\frac{ds}{dt}(\nabla f.N)N - kfN\right]|C|ds$$

$$\frac{ds}{dt} = (kf - (\nabla f.N)N)$$

where N is the unit normal to the curve c and k is its curvature. This modification improves the segmentation from incorrect regions and boundary discontinuities. However, this model relies on non-parameterized curve, and evolves an initial curve according to the boundary attraction term towards one direction (inwards/ outwards). Further, it demands a specific initialization step, where the initial curve should be completely exterior or interior to the real object boundaries. Lankton and Tannenbaum [211] and Shi and Karl [212] have addressed the problem of segmenting heterogeneous features in the image by reformulating region-based segmentation energy in a local way. Localized contours are capable of segmenting objects with heterogeneous features. However, the region based energy method gives very small value at the boundary and makes the speed of the moving contour low. D. Jayadevappa et al. [213] has proposed a modified variation of level set offering a long range attraction to overcome the drawbacks in [211, 212]. This method proposes a new speed term embedded with an edge indicator function in the evolution equation, which optimizes the effective distance of the attracting force and also provides robust edge estimation so that it can stop the contour even at weak or blurred edges. By introducing this function, the leakage problem is avoided effectively in most cases and it also captures range, which are improved compared to previous methods.

Geometric deformable models are based on the theory of curve evolution and are implemented using the level sets by Osher et al. 1988[214] numerical method. In particular, curves and surfaces are evolved using only geometric measures, resulting in an evolution that is independent of the parameterization. The mathematical form of level sets scheme is given as: $\partial\phi/\partial t$ = V (k)$|\nabla\phi|$ where V (k) is called speed function, k is curvature and $\phi$ is the level sets function. The topological adaptation can be useful in many applications, but it sometimes lead to undesirable results. Geometric deformable models, when applied to noisy images with ill-defined boundary, may produce shapes that have inconsistent topology with respect to the actual object by Xu et al., 2000 [215]. In these cases, the significance of ensuring a correct topology is often a necessary condition for many subsequent applications, while in the case of tumor segmentation it is very difficult or impossible. The topological adaptation can be useful in many applications, but it sometimes lead to undesirable results. Geometric deformable models, when applied to noisy images with ill-defined boundary, may produce shapes that have inconsistent topology with respect to the actual object, in the case of tumor segmentation it is very difficult or impossible.

**7.26 Hidden Markov Model:** To produce ever finer resolution in spectral, spatial and temporal data, Non-brain structures removed and it estimates the tissue intensity variation [216,217, 218, 219,220]. The formal specifications of HMM are completely characterized by the some general model parameters such as ; a) *N*, the number of states in the model, b) *M*, the number of mixtures in the random function, c) *A*, the state transition probability distribution. *A* = {$a_{ij}$}, where $a_{ij}$ is the transition probability from state *i* to state *j* . That is

$$a_{ij} = P[q_{t+1} = j | q_t = i], 1 \leq i, j \leq N$$

d) π, the initial state distribution, $\pi = \{\pi_i\}$ Where

$$\pi_i = P[q_i = i], 1 \leq i \leq N$$

This HMM topology finds good as well as its model parameters. This method is able to find the optimal states in all cases. The HMM can trained by genetic algorithms or more optimized methods. Thus we can use HMM as training data set.

**7.29 Genetic algorithms based:** A genetic algorithms are population based process to find exact or approximate solution to optimization the search problem is inspired by the generic process of biological organism used in computing. Genetic algorithms mainly consider by S. Chabrier et al. [221] in 2008 genotype, initial population, fitness function, operators on genotypes, stopping criterion. Individual description of the class of each pixel of the image segmentation results is called Genotype and initial population is the set of individuals' characterization by their genotypes. A fitness function is a particular type of function that is used to summarise and enables us to quantify the fitness of an individual to the environment by considering its genotype as a single figure of merit. K. Selvanayaki et al. [222]in 2012 use a fitness function $f(x) = 1/(1 + x^2)$
Where x is a window of size 5×5 and alterations on genotypes in order to make the population evolve during generations are described on the operation on genotypes. Mainly three types of operation are used, they are *i)* individual mutation or Mutation: A genetic operator that randomly selects a subset and replaces it with another randomly-created one. Its aim is to improve diversity as well as prevent being trapped in a local optimal solution. Individual's genes are modified in order to be better adapted to the environment. It use the no uniform mutation process which randomly selects one chromosome $x_i$, and sets it as equal to a no uniform random number [221]

$$x_i' = \begin{cases} x_i + (b_i - x_i)f(G) & \text{if } r_1 < 0.5 \\ x_i - (x_i + a_i)f(G) & \text{if } r_1 \geq 0.5 \end{cases}$$

Where

$$f(G) = \left(r_2\left(1 - G/G_{max}\right)\right)^b$$

$r_1$, $r_2$ are numbers in the interval [0, 1], $a_i$, $b_i$ are lower and upper bound of chromosome $x_i$, $G$ is the current generation, $G_{max}$ is the maximum number of generations, $b$ is a shape parameter. *ii)* Selection of an individual or Reproduction: A genetic operator that copies the individuals with the best fitness values directly into the population of the next generation without going through the crossover operation. Individuals that are not adapted to the environment do not survive to the next generation. The normalized geometric ranking selection method which defines a probability $P_i$ for each individual to be selected as follows:

$$P_i = \frac{q(1-q)^{r-1}}{1-(1-q)^n}$$

Where $q$ is the probability of selecting the best individual, $r$ is the rank of individual, $n$ is the size of the population. *iii)* Crossing-over: A genetic operator that randomly selects a sub-tree and replaces it with another randomly-created one. Its aim is to improve diversity as well as prevent being trapped in a local optimal solution. Two individuals can reproduce by combining their genes. The arithmetic crossover which produces two complementary linear combinations of the parents;

$$X' = aX + (1-a)Y$$
$$Y' = (1-a)X + aY$$

Where $X,Y$ are genotype of parents, $a$ is a number in the interval [0, 1], $X'$ and $Y'$ are genotype of the linear combinations of the parents. In Termination criterion execute the processes of fitness computation, selection, crossover, and mutation for a predetermined number of iterations. In every generational cycle, the fittest chromosome till the last generation is preserved – elitism [223, 224]. Thus on termination, this chromosome gives us the best solution encountered during the search. This criterion allows stopping the evolution of the population. It can consider the stability of the standard deviation of the evaluation criterion of the population or set a maximal number of iterations. Basically these five information [225, 226] execute of the genetic algorithm is carried out in four steps: (1) definition of the initial population (segmentation results) and computation of the fitness function (evaluation criterion) of each individual, (2) mutation and crossing-over of individuals, (3) selection of individuals, (4) evaluation of individuals in the population, (5) back to Step 2 if the stopping criterion is not satisfied.

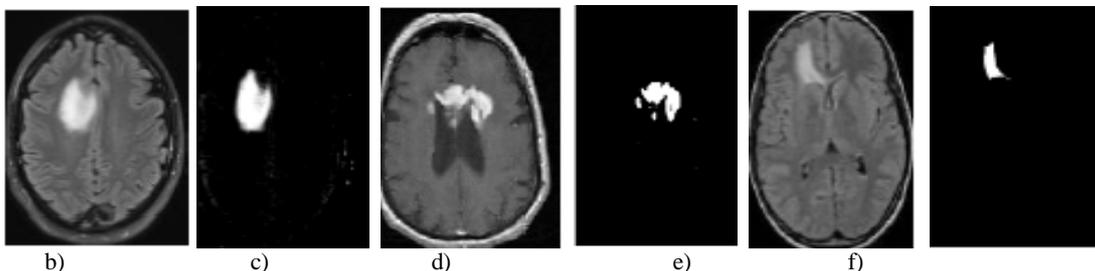

a)    b)    c)    d)    e)    f)

***Figure 30:*** *Output cited by Garima Garg (2012) [227] et. al. tumor segmentation using genetic algorithms (a, c, e) original image and (b, d, f) genetic based segmented image.*

Genetic Algorithm creates a sequence of populations for each successive generation by using a selection mechanism and uses operators such as crossover and mutation as principal search mechanisms - the aim of the algorithm being to optimize a given objective or fitness function. A prominent feature of MRI brain images is the fact that the texture patterns of the various tissues are fixed. Texture descriptors can be used to capture the salient features of the texture pattern in order to distinguish one type of tissue from another. Thus it can help refine the tumorous region already outlined by the existing process. GA is applied to enhance the detected border. The figure of merit is calculated to identify whether the detected border is exact or not. One of the disadvantages of the genetic algorithms is that it truly depends upon the fitness function.

**8. SUMMARY AND CONCLUSIONS:** Here several existing brain tumor segmentation and detection methodology has been discussed for MRI of brain image. All the steps for detecting brain tumor have been discussed including pre-processing steps. Pre-processing involves several operations like non local, Analytic correction methods, Markov random field methods and wavelet based methods has been discussed. Quality enhancement and filtering are important because edge sharpening, enhancement, noise removal and undesirable background removal are improved the image quality as well as the detection procedure. Among the different filtering technique discussed above, median filter suppressed the noise without blurring the edges and it is better outlier without reducing sharpness of the images, mean filter are much greater sensitive than that of median filter in the context of smoothing the image. Gaussian reduces the noise; enhance the image quality and computationally more efficient than other filtering methodology. After the several image quality improvement and noise reduction discussion here, some possible segmentation methodology like intensity based binarized segmentation, Region based, classification based, texture based, clustered based, neural network based, fuzzy, edge based, atlas, knowledge based, fusion, probabilistic segmentation has been described above with short description, advantage and disadvantage to detect or segment a brain tumor from MRI of brain image. In the threshold intensity based binnarized segmentation Kapur method is best methods and produce very effective results. Most of the binarized fails due to large intensity difference of foreground and background i.e. the black background of MRI image. In region growing methodologies are not standard methods for validate segmentation; the main problem is quality of segmentation in the border of tumor. This methods are good for homogeneous tumor but not for heterogeneous tumor. Classification based segmentation segment tumor accurately and produce good results for large data set but undesirable behaviours can occurs in case where a class is under represented in training data. Clustered based segmentation performs very simple, fast and produce good results for non-noise image but for noise images it leads to serious inaccuracy in the segmentation. In neural network based segmentation perform little better on noise field and no need of assumption of any fundamental data allocation but learning process is one of the great disadvantages of it. In edge based segmentation canny perform well but other noise edge are also produce, sobel gives very effective results but disadvantage of its are the discontinuity of edges. In the case of a weak border or a gap in the border of the tumor, the contour or surface may leak to other regions. In spite of several disadvantage, an automization of brain tumor segmentation using combination of threshold based and classification like SVM, Basian may overcome the problems and gives effective and accurate results for brain tumor detection.